\documentclass[preprint,12pt]{elsarticle}

\usepackage{amssymb}
\usepackage{amssymb}
\usepackage{float}
\usepackage {amsmath}
\usepackage{algorithm}  
\usepackage{algpseudocode}  
\usepackage{amsmath}  
\usepackage{booktabs} 
\usepackage{caption} 
\usepackage{setspace}
\usepackage{subcaption}
\usepackage{hyperref}
\usepackage{threeparttable}
\usepackage{multirow}
\usepackage{doi}
\usepackage{xcolor} 
\usepackage{caption}
\usepackage{mathrsfs}
\usepackage{bm}
\usepackage{adjustbox}
\usepackage{amsfonts}

\hypersetup{
    colorlinks=true, 
    linkcolor=blue,  
    urlcolor=blue,   
    citecolor=blue   
}

\makeatletter
\renewcommand{\fnum@figure}{\textbf{Fig. \thefigure.}\@gobble}
\makeatother

\makeatletter
\renewcommand{\fnum@table}{\textbf{Table \thetable }\@gobble}
\makeatother

\journal{Elsevier}

\begin{document}

\vspace{30pt}

\noindent \textbf{Title: }

\noindent \textbf{Open-Set Fault Diagnosis in Multimode Processes via Fine-Grained Deep Feature Representation}
\\
\\
\noindent \textbf{Names of authors: }

\noindent Guangqiang Li $^a$, M. Amine Atoui $^b$, Xiangshun Li $^{a*}$ 
\\
\\
\noindent \textbf{Affiliations and addresses:}

\noindent $^a$ School of Automation, Wuhan University of Technology,  Wuhan, China

\noindent $^b$ School of Information Technology, Halmstad University,  Halmstad, Sweden
\\
\\
\noindent \textbf{Corresponding author}

\noindent Name: Xiangshun Li

\noindent E-mail: lixiangshun@whut.edu.cn

\begin{frontmatter}

\title{
Open-Set Fault Diagnosis in Multimode Processes via Fine-Grained Deep Feature Representation
}

\author[a]{Guangqiang Li} 
\ead{guangqiangli@whut.edu.cn}

\author[b]{M. Amine Atoui} 
\ead{amine.atoui@gmail.com}

\author[a]{Xiangshun Li\corref{cor1}} 
\ead{lixiangshun@whut.edu.cn}
\cortext[cor1]{Corresponding author.}

\affiliation[a]{organization={School of Automation},
            addressline={Wuhan University of Technology}, 
            city={Wuhan},
            postcode={430070}, 
            country={PR China}}
\affiliation[b]{organization={The School of Information Technology},
            addressline={Halmstad University}, 
            city={Halmstad},
            country={Sweden}}

\begin{abstract}
A reliable fault diagnosis system should not only accurately classify known health states but also effectively identify unknown faults. In multimode processes, samples belonging to the same health state often show multiple cluster distributions, making it difficult to construct compact and accurate decision boundaries for that state. 
To address this challenge, a novel open-set fault diagnosis model named fine-grained clustering and rejection network (FGCRN) is proposed. 
It combines multiscale depthwise convolution, bidirectional gated recurrent unit and temporal attention mechanism to capture discriminative features. 
A distance-based loss function is designed to enhance the intra-class compactness. 
Fine-grained feature representations are constructed through unsupervised learning to uncover the intrinsic structures of each health state. 
Extreme value theory is employed to model the distance between sample features and their corresponding fine-grained representations, enabling effective identification of unknown faults. 
Extensive experiments demonstrate the superior performance of the proposed method.

\end{abstract}



\begin{keyword}
Fault Diagnosis \sep Open set \sep Multimode processes
\end{keyword}

\end{frontmatter}

\section{Introduction}
\label{sec1}

Safety is the fundamental prerequisite for modern industrial production and manufacturing. 
With the growing scale and complexity of modern industrial systems, the probability of system faults has risen significantly \cite{RN266839,RN246049,RN266866}. 
Local faults may propagate within the system, potentially triggering cascading failures and catastrophic accidents, causing serious economic losses and threatening human safety \cite{RN260370,RN266841}. 
Fault diagnosis models aim to identify anomalies and assess system health state in real time, thereby enabling early warning and maintenance actions \cite{RN266842}. 
This timely intervention can effectively prevent fault propagation and escalation. 
Therefore, fault diagnosis models serve as the “doctor” of industrial systems, providing critical technical support for their long-term safe, stable, and reliable operation. 

With the development of advanced sensing, communication, and storage technologies, industrial systems have accumulated a large amount historical data containing important diagnostic information \cite{RN266863}. 
Based on the collected data, data-driven methods aim to learn fault-related features to accurately identify health state categories.
Compared to model-based and knowledge-based methods, data-driven methods have significant strengths as they do not rely on precise mathematical models and extensive expert knowledge \cite{RN266844}. 
As a representative of data-driven methods, machine learning methods have obtained widespread attention in fault diagnosis and achieved promising outcomes. 
Traditional machine learning methods such as support vector machine (SVM) \cite{RN266845}, Bayesian network \cite{RN266847}, discriminant analysis \cite{RN266848,RN266865}, and random forest \cite{RN266849} have been extensively employed. 
As industrial systems become increasingly complex, the extraction of fault features has become increasingly challenging. 
As an important branch of machine learning, deep learning has gained increasing interest for its great capability in automatic feature extraction and high diagnostic accuracy \cite{RN266846,RN266850}. 
Various network architectures, including convolution neural network \cite{RN260387}, long short-term memory \cite{RN266851,RN266852}, transformer \cite{RN266854}, auto-encoder \cite{RN266855,RN266856}, generative adversarial network \cite{RN266858,RN266857}, and prototypical network \cite{RN266864}, have been employed in fault diagnosis. 
However, most existing studies assume that industrial systems operate under a single working condition. 

Due to changes in environment conditions and production strategies, industrial plants frequently switch between different operating modes. 
The coupling between operating modes and fault types significantly increases the complexity of fault diagnosis in multimode conditions. 
Li et al. \cite{RN266859} employed instance normalization to suppress mode-specific features and constructed a temporal attention to attend to critical moments with high mode-invariant information. 
Qin et al. \cite{RN266426} designed the adaptive attention mechanism to enhance critical features and introduced a triplet loss to improve discriminative performance under multimode conditions. 
Wu et al. \cite{RN260394} combined fine-tuning with the joint adaptation network to facilitate fault knowledge transfer across different modes. 
Yang et al. \cite{RN243235} used adversarial training to effectively capture shared representations by minimizing the Wasserstein distance between different modes, thereby improving classification performance. 

Deep learning-based methods typically rely on the assumption that the training and test datasets have the same label space, but this is usually difficult to satisfy in practical systems. 
The historical data collected for training usually fail to cover all possible fault categories, especially rare and infrequent faults. 
In this context, effectively classifying known health states and identifying unknown faults is crucial for ensuring reliable decision-making and operational management \cite{lou2022novel}. 
Yu et al. \cite{RN261923} employed extreme value theory to identify unknown faults, while Peng et al. \cite{RN262258} introduced the soft Brownian offset to generate synthetic samples for unknown fault identification.
Out-of-distribution (OOD) detection and open-set recognition in computer vision are highly similar to unknown fault identification in fault diagnosis and have been extensively studied. 
Bendale et al. \cite{RN266834} proposed OpenMax, a method that estimates the probability of a sample belonging to an unknown category based on the distance between its activation vector and class activation vectors. 
Additionally, various scoring functions have been proposed to distinguish between known and unknown categories, including maximum softmax probability (MSP) \cite{RN266833}, maximum logit (MaxLogit) \cite{RN266836}, KL matching score \cite{RN266836}, generalized entropy (GEN) score \cite{RN266837} and virtual-logit matching (ViM) \cite{RN266838} score.

However, identifying unknown faults in multimode processes presents significant challenges. 
First, samples from the same category often exhibit a multicluster structure in the feature space. 
This discontinuity weakens the performance of threshold-based open-set recognition methods that rely on distance, probability, or confidence scores.
Second, the significant intra-class variation further hinders the construction of compact, closed and continuous decision boundaries that strictly enclose samples of a single category, thereby increasing the risk of misidentification.

\begin{figure}[!ht]
\centerline{\includegraphics[width=1\columnwidth,]{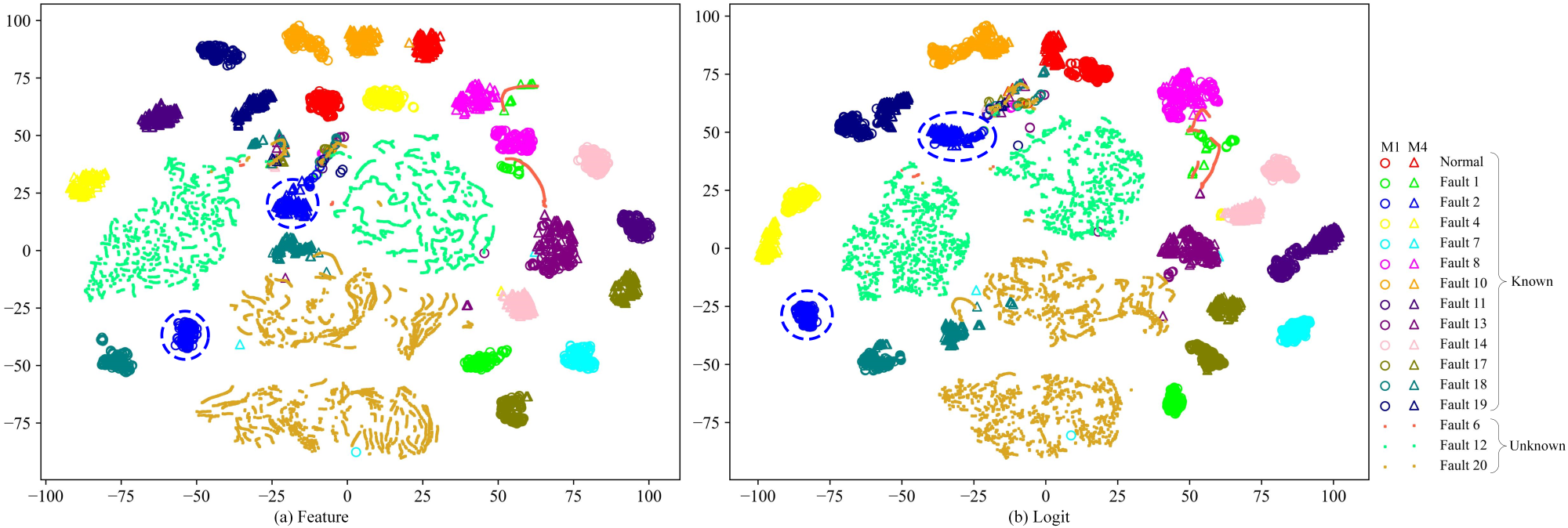}}
	\caption{Visualization of the TE process.}
	\label{Fig1}
\end{figure}

For instance, \hyperref[Fig1]{Fig. 1} shows the t-SNE visualizations of the features and Logit outputs generated by the proposed model under operating modes 1 and 4 of the Tennessee Eastman (TE) process dataset. 
The hollow circles and hollow triangles represent the samples from modes 1 and 4, respectively.
The dashed region highlights the samples corresponding to Fault 2. 
It is evident that these samples do not form a single cohesive cluster in either the feature or Logit space. 
Instead, they appear to be divided into two sub-clusters corresponding to different operating modes. 
This discontinuity in both spaces makes it difficult to define a reliable threshold for unknown fault identification and to construct a continuous decision boundary that only encloses the Fault 2 category. 

To address the above problems, a fine-grained clustering and rejection network
(FGCRN) is proposed. 
Unlike traditional methods that assign a single feature representation to each health state, this work constructs multiple representations for each state. 
Deep discriminative features are extracted using a combination of multiscale depthwise convolution (MSDC), bidirectional gated recurrent unit (BiGRU), and temporal attention mechanism (TAM), while a distance-based loss is incorporated to enhance feature compactness. 
Furthermore, unsupervised clustering is applied to divide each health state into multiple sub-clusters, using the cluster centroids as fine-grained representations. 
Based on extreme value theory (EVT), the rejection probability is estimated using the Mahalanobis distance between the sample and its corresponding cluster centroid, thereby achieving effective identification of unknown faults.

The main contributions of this paper are detailed as follows:

(1) A novel fine-grained clustering and rejection network is proposed for open-set fault diagnosis in multimode processes. 
It is capable of effectively classifying known health states and accurately identifying unknown faults. 

(2) Multiscale depthwise convolution, bidirectional gated recurrent unit, and temporal attention mechanism are integrated to capture key temporal features, while the combination of batch normalization and self-adaptive instance normalization adaptively preserves discriminative statistical information, thereby facilitating accurate classification of known health states. 

(3) Unsupervised clustering is employed to construct fine-grained category representations, and combined with extreme value theory to achieve effective identification of unknown faults. 

(4) Extensive experiments conducted on two simulated datasets and an actual industrial process dataset demonstrate that the proposed method outperforms existing advanced methods in identifying unknown faults.

The remaining sections of this paper are structured as follows. 
\hyperref[sec2]{Section 2} covers the problem formulation, the basics of clustering methods, and the application of extreme value theory in unknown fault identification.
\hyperref[sec3]{Section 3} details the construction and optimization process of the proposed method, and \hyperref[sec4]{Section 4} validates its performance across multiple datasets. 
\hyperref[sec5]{Section 5} outlines the main conclusions.

\section{Preliminaries}
\label{sec2}

\subsection{Problem formulation}

Let the training set be denoted as $D_\mathrm{tr}=\{ x_\mathrm{tr}^{i}, y_\mathrm{tr}^{i} \}_{i=1}^{N_\mathrm{tr}}$, containing $N_\mathrm{tr}$ samples collected under $M$ different operating modes. 
Each sample $x_\mathrm{tr}^{i} \in \mathbb{R}^{V \times T}$ represents a multivariate time window with $V$ monitored variables and a window length of $T$; $y_\mathrm{tr}^{i} \in Y_\mathrm{tr} = \{Y_1, Y_2, \dots, Y_k\}$ denotes its health state label. 
The label set consists of one normal (healthy) category and $k-1$ known fault categories. 
The test set is denoted as $D_\mathrm{te} = \{ x_\mathrm{te}^{i} \}_{i=1}^{N_\mathrm{te}}$, also collected under the same $M$ operating modes.
However, its label set expands to $Y_\mathrm{te} = \{Y_1, Y_2, \dots, Y_k, Y_{k+1}, \dots, Y_{k+u}\}$, where the additional $u$ categories represent unknown faults not seen during training. 
Note that neither the training set nor the test set includes operating mode labels.
The objective of open-set fault diagnosis in this context is to learn a model $H$ from $D_\mathrm{tr}$ that can not only accurately classify samples from the known health state categories $\{ Y_1, Y_2, \dots, Y_k \}$ but also reliably identify and reject samples from the unknown fault categories $\{ Y_{k+1}, \dots, Y_{k+u} \}$, as unseen or novel.

\subsection{K-means++ }
K-means++ \cite{arthur2007k} is an unsupervised clustering model designed to automatically capture latent group structures in unlabeled data, ensuring that intra-cluster similarity is higher than inter-cluster similarity. 
It assigns each sample to the nearest cluster based on distance metrics such as Euclidean or Mahalanobis distance, and iteratively updates each cluster center as the average of the samples within the cluster until convergence. 
Compared to the K-means model, which initializes cluster centers randomly and may yield unstable results, K-means++ improves initialization by selecting centers that are maximally distant from each other. 
This strategy increases the likelihood that the initial cluster centers locate in different actual clusters, thereby improving the stability of the clustering results. 
The detailed steps of the K-means++ cluster process are presented in \hyperref[alg1]{Algorithm 1}.

\begin{algorithm}[!ht]  
	\caption{K-means++.}  
	\label{alg1}  
1: Randomly select one sample as the first cluster center. \\
2: Calculate the minimum distance $d_i$ between each sample and the existing cluster centers.\\
3: Compute the probability of selecting each sample $d_i/\sum_k d_k$.\\
4: Choose the next cluster center based on these calculated probabilities. \\
5: Repeat Steps 2--4 until $K$ cluster centers are obtained.\\
6: Assign each sample to the nearest cluster based on the distances.\\
7: Update the cluster centers using the mean of the samples within each cluster.\\
8: Repeat Steps 6 and 7 until the cluster centers converge. 
\end{algorithm}  

\subsection{Extreme value theory for unknown fault identification}
\begin{figure}[!ht]
\centerline{\includegraphics[width=0.8\columnwidth,]{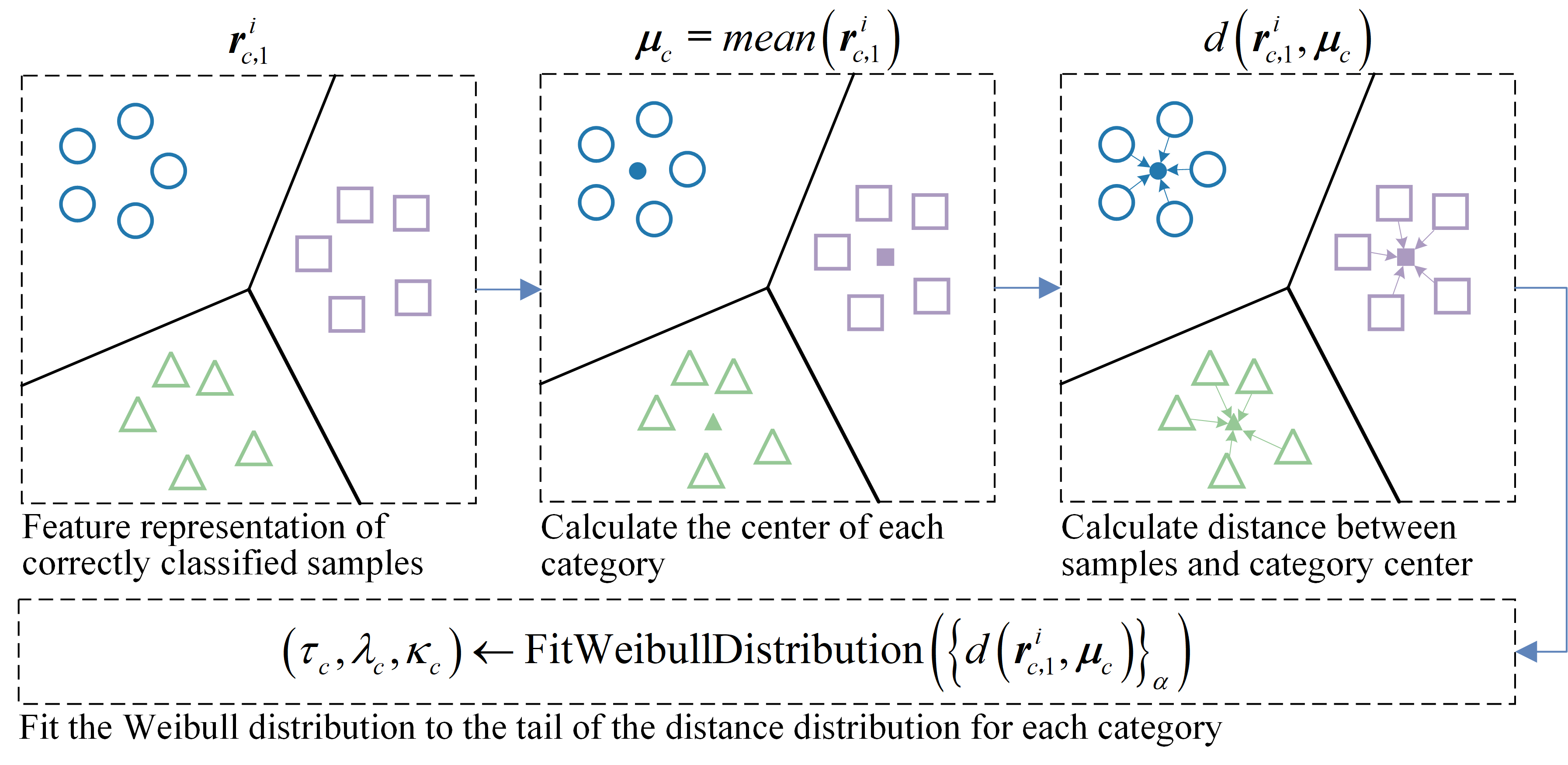}}
	\caption{EVT Modeling for unknown fault identification.}
	\label{Fig2}
\end{figure}

Extreme value theory (EVT) is an effective method for analyzing the distribution of abnormally high or low values \cite{RN266861,RN266862}. 
It estimates the probability of a sample belonging to the unknown category by modeling the distance distribution between samples and the center of its corresponding known category\cite{RN266860}. 
The overall procedure is illustrated in \hyperref[Fig2]{Fig. 2}.
First, for each known health state, the mean feature vector of correctly classified samples is taken as the category representation. 
Subsequently, the distances between each correctly classified sample and its corresponding category representation are calculated. 
The tail of this distance distribution $ \left\{d\left(\boldsymbol{r}_{c, 1}^{i}, \boldsymbol{\mu}_{c}\right)\right\}_{\alpha} $, defined as the top $ \alpha $ largest distances, is fitted using a Weibull distribution. 
The rejection probability for each sample is given by the cumulative distribution function of the Weibull model.  
During testing, samples with rejection probabilities exceeding the predefined threshold are identified as the unknown fault.

\section{Proposed method}
\label{sec3}

\subsection{Overall architecture of FGCRN}
\begin{figure}[!ht]
\centerline{\includegraphics[width=1\columnwidth,]{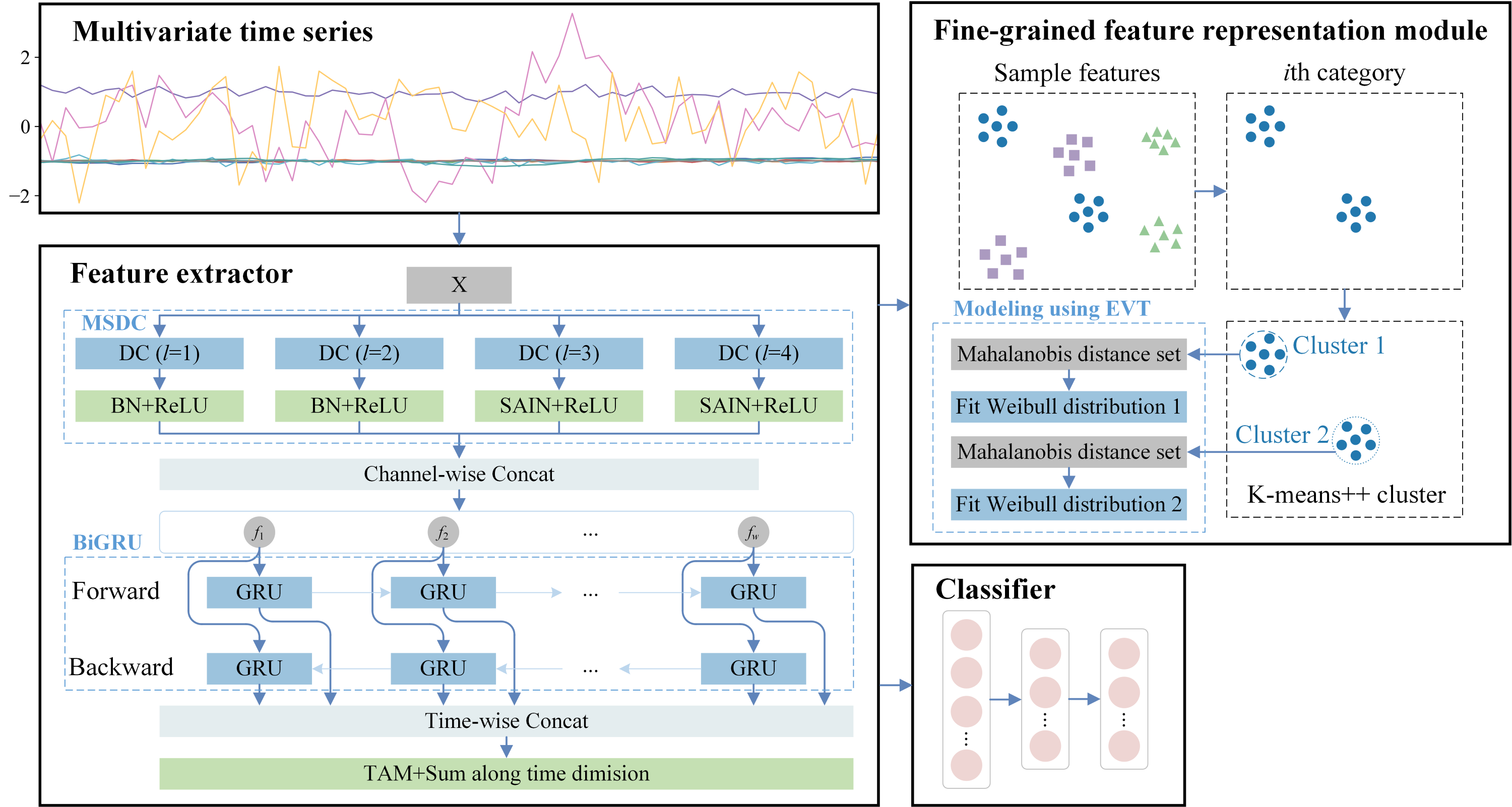}}
	\caption{Overall architecture of  FGCRN.}
	\label{Fig3}
\end{figure}

The overall architecture of the proposed model is illustrated in \hyperref[Fig3]{Fig. 3}. 
It includes a feature extractor, a classifier, and a fine-grained feature representation module. 
The feature extractor captures discriminative features closely associated with the health states. 
The classifier determines the health state category based on the extracted features. The fine-grained feature representation module divides the feature space of each health state based on distance differences. EVT is applied to model the known categories, enabling the identification of previously unseen faults.

\subsection{Feature extractor}

The feature extractor is composed of three main components: a MSDC to extract local features across multiple scales, a BiGRU to capture temporal dependencies in both forward and backward directions, and a TAM to enhance the representation of critical time points.  

\subsubsection{MSDC}
The MSDC employs four kernel sizes to extract local features across different temporal scales. 
Each channel is processed by an independent convolution kernel, which effectively reduces computational complexity. 
The convolution output at the $j$th channel and $t$th time step is formulated as, 

\begin{equation}
	\begin{aligned}
		f_{t,j}^{\mathrm{DC}_{2l+1}}=\sum_{k=t-l}^{t+l}x_{k,j}w_{k+l-t+1,j}, \quad l=1,2,3,4,
	\end{aligned}	
	\label{Eq1}	
\end{equation}
where $w$ denotes the convolution kernel and the kernel size is $2l+1$. 

Batch normalization (BN) is applied after depthwise convolutions with kernel sizes of 3 and 5 to accelerate convergence. 
The output features after BN processing are formulated as,
\begin{equation}
	\begin{aligned}
		f_{t, j}^{\mathrm{BN}}&=\gamma_{j}^{\mathrm{BN}}\left(\frac{f_{t, j}^{\mathrm{DC}}-\mu_{j}^{\mathrm{BN}}}{\sqrt{\left(\sigma_{j}^{\mathrm{BN}}\right)^{2}+\varepsilon}}\right)+\beta_{j}^{\mathrm{BN}}, \\
\mu_{j}^{\mathrm{BN}}&=\frac{1}{B T} \sum_{b=1}^{B} \sum_{t=1}^{T} f_{t, j}^{\mathrm{DC}}, \\
\sigma_{j}^{\mathrm{BN}}&=\sqrt{\frac{1}{B T} \sum_{b=1}^{B} \sum_{t=1}^{T}\left(f_{t, j}^{\mathrm{DC}}-\mu_{j}^{\mathrm{DC}}\right)^{2}},
	\end{aligned}	
	\label{Eq2}	
\end{equation}
where $B$ denotes batch size, $\gamma_{j}^{\mathrm{BN}}$ and $\beta_{j}^{\mathrm{BN}}$ are learnable weights. 
$\mu_{j}^{\mathrm{BN}}$ and $\sigma_{j}^{\mathrm{BN}}$ denote the mean and standard deviation of the $j$th channel within a batch.

For kernel sizes of 7 and 9, self-adaptive instance normalization (SAIN) \cite{li2025fault} is employed to weaken statistical features unrelated to the healthy state category. 
The output features after SAIN processing are formulated as,
\begin{equation}
	\begin{aligned}
		  f_{t,j}^{\mathrm{SAIN}}&=\gamma_{j}^{\mathrm{SAIN}}\left(\frac{f_{t,j}^{\mathrm{DC}}-\mu_{j}^{\mathrm{SAIN}}}{\sqrt{\left(\sigma_{j}^{\mathrm{SAIN}}\right)^{2}+\varepsilon}}\right)+\beta_{j}^{\mathrm{SAIN}}, \\
 \mu_{j}^{\mathrm{SAIN}}&=\frac{1}{T}\sum_{t=1}^{T}f_{t,j}^{\mathrm{DC}}, \\
  \sigma_{j}^{\mathrm{DC}}&=\sqrt{\frac{1}{T}\sum_{t=1}^{T}\left(f_{t,j}^{\mathrm{DC}}-\mu_{j}^{\mathrm{DC}}\right)^{2}}, \\
 \gamma^{\mathrm{SAIN}}&=g_2\left(\mathrm{ReLU}\left(g_1\left(\mu_j^{\mathrm{SAIN}}\right)\right)\right), \\
  \beta^{\mathrm{SAIN}}&=g_{4}\left(\mathrm{ReLU}\left(g_{3}\left(\sigma_{j}^{\mathrm{SAIN}}\right)\right)\right),
	\end{aligned}	
	\label{Eq3}	
\end{equation}
where $g1,g2,g3$ and $g4$ denote fully connected layers, $\mu_{j}^{\mathrm{SAIN}}$ and $\sigma_{j}^{\mathrm{SAIN}}$ denote the mean and standard deviation of each sample in the $j$th channel. 

The combination of BN and SAIN contributes to enhancing feature diversity. 
SAIN adaptively suppresses statistical interference, while BN preserves statistical discriminative information. 
By concatenating the outputs of both operations along the channel dimension, the network is enabled to automatically select the most discriminative features for classification. 
The concatenated feature is formulated as, 
\begin{equation}
	\begin{aligned}
f_{t}=\mathrm{ReLU}\left(\left[f_{t}^{\mathrm{BN}_{3}};f_{t}^{\mathrm{BN}_{5}};f_{t}^{\mathrm{SAIN}_{7}};f_{t}^{\mathrm{SAIN}_{9}}\right]_{\mathrm{C}}\right),
	\end{aligned}	
	\label{Eq4}	
\end{equation}
where $\left[ \quad \right]_{\mathrm{C}}$ denotes concatenation along the channel dimension, and $f_{t}\in\mathbb{R}^{4V}$. 

\subsubsection{BiGRU}
\begin{figure}[!ht]
\centerline{\includegraphics[width=1.\columnwidth,]{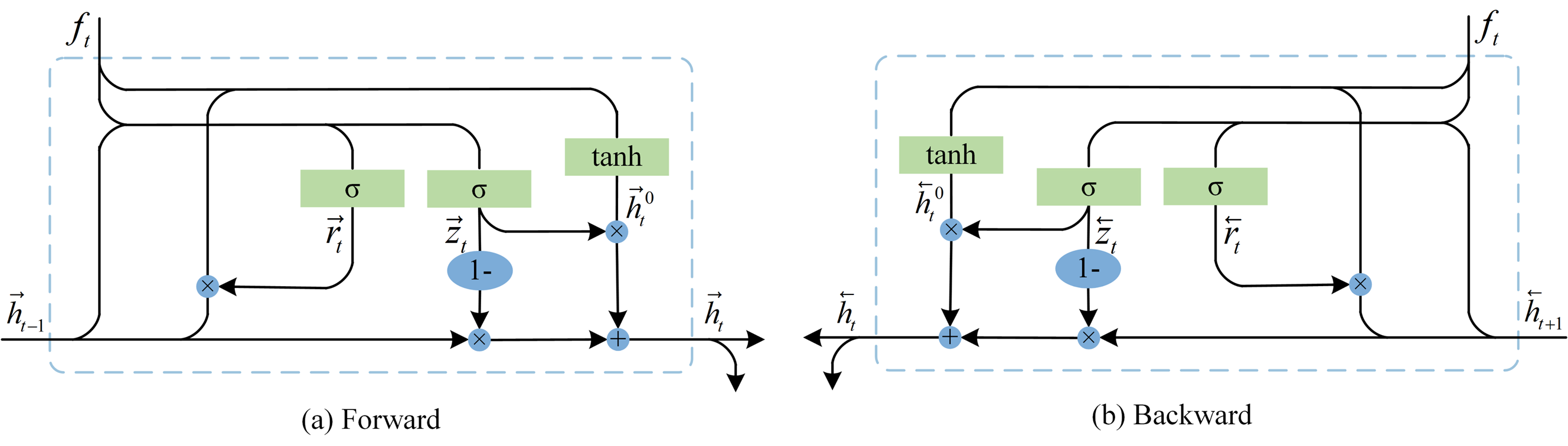}}
	\caption{Forward and backward flows of BiGRU.}
	\label{Fig4}
\end{figure}

As an important variant of recurrent neural network, GRU transmits information through hidden states and are superior to CNN in capturing long-term dependencies. 
The internal information flow of the GRU is illustrated in \hyperref[Fig4]{Fig. 4 (a)}. 
The hidden state at the current time step $h_{t}$ is determined by the previous hidden state $h_{t-1}$ and the current input feature $f_{t}$. 
This process is formulated as,

\begin{equation}
	\begin{aligned}
 & \overrightarrow{r}_{t}=\sigma\left(\overrightarrow{W}_rf_t+\overrightarrow{U}_r\overrightarrow{h}_{t-1}\right), \\
 & \overrightarrow{h}_{t}^0=\tanh\left(\overrightarrow{W}f_t+\overrightarrow{U}\left(\overrightarrow{r}_t\odot \overrightarrow{h}_{t-1}\right)\right), \\
 & \overrightarrow{z}_{t}=\sigma\left(\overrightarrow{W}_zf_t+\overrightarrow{U}_z\overrightarrow{h}_{t-1}\right), \\
 & \overrightarrow{h}_{t}=(1-\overrightarrow{z}_t)\odot \overrightarrow{h}_{t-1}+\overrightarrow{z}_t\odot\overrightarrow{h}_t^0,
\end{aligned}
	\label{Eq5}	
\end{equation}
where $\overrightarrow{r}_{t}$ and $\overrightarrow{z}_{t}$ control information reset and update, respectively. 
$\sigma$ is the sigmoid activation function; $ \overrightarrow{W}_r, \overrightarrow{W}_z, \overrightarrow{W}, \overrightarrow{U}_r, \overrightarrow{U}_z$ and $\overrightarrow{U}$ are weight matrices. 
Compared to GRU, BiGRU combines forward and backward flows, enabling a more comprehensive capture of temporal context information in sequence data. 
The forward output of BiGRU is given in \hyperref[Eq5]{Eq. 5}, and the backward output is formulated as,

\begin{equation}
	\begin{aligned}
 & \overleftarrow{r}_{t}=\sigma\left(\overleftarrow{W}_rf_t+\overleftarrow{U}_r\overleftarrow{h}_{t+1}\right), \\
 & \overleftarrow{h}_{t}^0=\tanh\left(\overleftarrow{W}f_t+\overleftarrow{U}\left(\overleftarrow{r}_t\odot \overleftarrow{h}_{t+1}\right)\right), \\
 & \overleftarrow{z}_{t}=\sigma\left(\overleftarrow{W}_zf_t+\overleftarrow{U}_z\overleftarrow{h}_{t+1}\right), \\
 & \overleftarrow{h}_{t}=(1-\overleftarrow{z}_t)\odot \overleftarrow{h}_{t+1}+\overleftarrow{z}_t\odot\overleftarrow{h}_t^0.
\end{aligned}
	\label{Eq6}	
\end{equation}
As shown in \hyperref[Fig4]{Fig. 4 (b)}, during the backward information flow, the hidden state at the current time step $\overleftarrow{h}_{t}$ is determined by the subsequent hidden state $\overleftarrow{h}_{t+1}$ and the current input feature $f_t$.

Finally, the forward and backward outputs are concatenated along the time dimension to fully preserve the context information. 
The resulting concatenated features are formulated as,
\begin{equation}
	\begin{aligned}
 h=\left[ \overrightarrow{h}_{1}, \ldots, \overrightarrow{h}_{T}, \overleftarrow{h}_{1}, \ldots, \overleftarrow{h}_{T}\right]_\mathrm{T},
\end{aligned}
	\label{Eq7}	
\end{equation}
where $\left[ \quad \right]_{\mathrm{T}}$ denotes concatenation along the time dimension. 

\subsubsection{TAM}
The forward and backward outputs of BiGRU contribute differently to health state classification. 
To enhance the model’s focus on critical moments enriched with discriminative contextual information, the TAM \cite{RN266859} is introduced. 
The temporal attention map is formulated as follows,
\begin{equation}
	\begin{aligned}
 & a_{t}=\mathrm{ReLU}\left(\mathrm{Conv}\left(\left[a_{1};a_{2}\right]_{\mathrm{C}}\right)\right), \\
 & a_{t}^1=g_6\left(\mathrm{ReLU}\left(g_5(\mathrm{Avg}\left(h_{t}\right))\right)\right), \\
 & a_{t}^2=g_8\left(\mathrm{ReLU}\left(g_7(\operatorname{\mathrm{Std}}\left(h_{t}\right)\right))\right),
\end{aligned}
	\label{Eq8}	
\end{equation}
where Avg and Std are average pooling and standard deviation operations; $g_5,g_6,g_7$ and $g_8$ are fully connected layers; Conv denotes convolution operation. 

The output features of the BiGRU are weighted by the temporal attention map to emphasize critical time steps, and subsequently aggregated along the time dimension. 
The final representation is formulated as,
\begin{equation}
	\begin{aligned}
r=\sum_{t=1}^{2T}h_{t}a_{t}. 
\end{aligned}
	\label{Eq9}	
\end{equation}

\subsection{Classifier}
The classifier maps the captured features $r$ to the health state predictions through a fully connected layer. 
The probability that the $i$th sample is assigned to the $c$th health state category is formulated as,
\begin{equation}
	\begin{aligned}
p_{c}^{i}&=\frac{\exp \left(o_{c}^{i}\right)}{\sum_{c=1}^{k} \exp \left(o_{c}^{i}\right)}, \\
o^{i}&=g_{9}\left(r^{i}\right),
\end{aligned}
	\label{Eq10}	
\end{equation}
where $g_9$ denotes the fully connected layer and $k$ denotes the number of known health state categories. 

\subsection{Fine-grained feature representation module}
In multimode processes, changes in operating conditions often lead to significant distribution differences among samples with the same health states. 
Constrained by the Softmax activation function, traditional fault diagnosis models can only classify samples into predefined health states. 
While this setup can distinguish known health states, it inevitably expands the feature space occupied by each state. 
As a result, the model learns coarse-grained feature representations, increasing the risk of misidentifying unknown faults as known health states. 
Motivated by this limitation, the fine-grained feature representations corresponding to each health state in different operating modes are modeled separately to improve the identification ability of unknown faults.
K-means++ is employed to cluster samples for each health condition and reveal their intrinsic grouping structure. It is applied here to partition the samples into potential $M$ modes (line 3 of \hyperref[alg2]{Algorithm 2}). 

Subsequently, the Mahalanobis distance between each correctly classified sample and its corresponding cluster center is calculated (line 6 of \hyperref[alg2]{Algorithm 2}). 
The distance is formulated as, 
\begin{equation}
	\begin{aligned}
    d\left(r_{c,1}^{i},\mu_{c}^m\right)=\sqrt{\mathrm{max}\{\left(r_{c,1}^{i}-\mu_{c}^m \right)^\mathrm{T} \left(\Sigma_{c}^{m}+\epsilon \mathrm{I}\right)^{-1} \left(r_{c,1}^{i}-\mu_{c}^m \right),d_0\}}, 
    \end{aligned}
    \label{Eq11}	
\end{equation}
where $r_{c,1}^{i}$ denotes the feature representation of the $i$th sample that is correctly classified into $c$th category, $\mu_{c}^m$ denotes the mean vector of the $m$th operating mode cluster within $c$th category, $\Sigma_{c}^{m}$ is the covariance matrix, $\epsilon \mathrm{I}$ is the regularization term, and $d_0$ denotes the minimum distance threshold. 

The tail distribution of Mahalanobis distances is then fitted using the Weibull distribution (line 7 of \hyperref[alg2]{Algorithm 2}), and its cumulative distribution function is used to estimate the rejection probability.
The rejection probability is formulated as,
\begin{equation}
	\begin{aligned}
    q_c^i=1-\exp\left(\frac{-\left\|d\left(r_{c,1}^i,\mu_c^m\right)-\tau_c^m\right\|}{\lambda_c^m}\right)^{\kappa_c^m}, 
    \end{aligned}
    \label{Eq12}	
\end{equation}
where $\tau_c^m$, $\lambda_c^m$ and $\kappa_c^m$ are the parameters of the Weibull distribution.

\begin{algorithm}[!ht]  
	\caption{Fine-grained feature representation modeling.}  
	\label{alg2}  
\textbf{Input: }Sample feature $r$, real category $y$, predicted category $\mathrm{argmax}\{ p \} $.\\
\textbf{Output: } Cluster center $\mu$, cluster covariance $\Sigma$, fitted Weibull distribution's parameters $\tau$, $\lambda$ and $\kappa$.\\
1: \textbf{for} $c$ = 1 to $k$ \textbf{do} \\
2: \ \ \ \ \ Select correctly classified sample features $r_{c,1}$. \\
3: \ \ \ \ \ Partition samples into $M$ clusters using the K-means++ algorithm.\\
4: \ \ \ \ \ \textbf{for} $m$ = 1 to $M$ \textbf{do} \\
5: \ \ \ \ \ \ \ \ \ Compute mean vector and covariance matrix for each cluster.\\
6: \ \ \ \ \ \ \ \ \ Compute the Mahalanobis distance between each sample and its corresponding cluster distribution via \hyperref[Eq11]{Eq. 11}.\\
7: \ \ \ \ \ \ \ \ \ Fit the Weibull distribution to the tail of the distance distribution and obtain the parameters $\tau_c^m$, $\lambda_c^m$ and $\kappa_c^m$.\\
8:\ \ \ \ \ \textbf{end for}\\
9: \textbf{end for}
\end{algorithm}  

\subsection{Optimization object}
The cross-entropy loss is employed to assess the model's classification performance on known categories and is formulated as,
\begin{equation}
	\begin{aligned}
    L_1=\sum_{i=1}^{N_\mathrm{tr}}\mathrm{log}\left({p_{y_\mathrm{tr}^i}}\right), 
    \end{aligned}
    \label{Eq13}	
\end{equation}
where $p_{y_\mathrm{tr}^i}$ denotes the $y_\mathrm{tr}^i$th output of $p$.
The distance loss is employed to enhance intra-class cohesion by minimizing the Mahalanobis distance between samples and their corresponding cluster centers, and is formulated as,
\begin{equation}
	\begin{aligned}
    L_2=\frac{1}{N_\mathrm{tr,1}}\sum_{i=1}^{N_\mathrm{tr,1}}\sqrt{\mathrm{max}\{\left(r_{c,1}^{i}-\mu_{c}^m \right)^\mathrm{T} \left(\Sigma_{c}^{m}+\epsilon \mathrm{I}\right)^{-1} \left(r_{c,1}^{i}-\mu_{c}^m \right),d_0\}}, 
    \end{aligned}
    \label{Eq14}	
\end{equation}
where $N_\mathrm{tr,1}$ denotes the number of correctly classified samples. 

The total loss is composed of the cross-entropy loss and the distance loss, and is formulated as, 
\begin{equation}
	\begin{aligned}
    L=L_1+\lambda L_2, 
    \end{aligned}
    \label{Eq15}	
\end{equation}
where $\lambda$ denotes the hyperparameter that balances the contributions of cross-entropy and distance losses.

\subsection{Application workflow of FGCRN}

\begin{figure}[!ht]
\centerline{\includegraphics[width=1\columnwidth,]{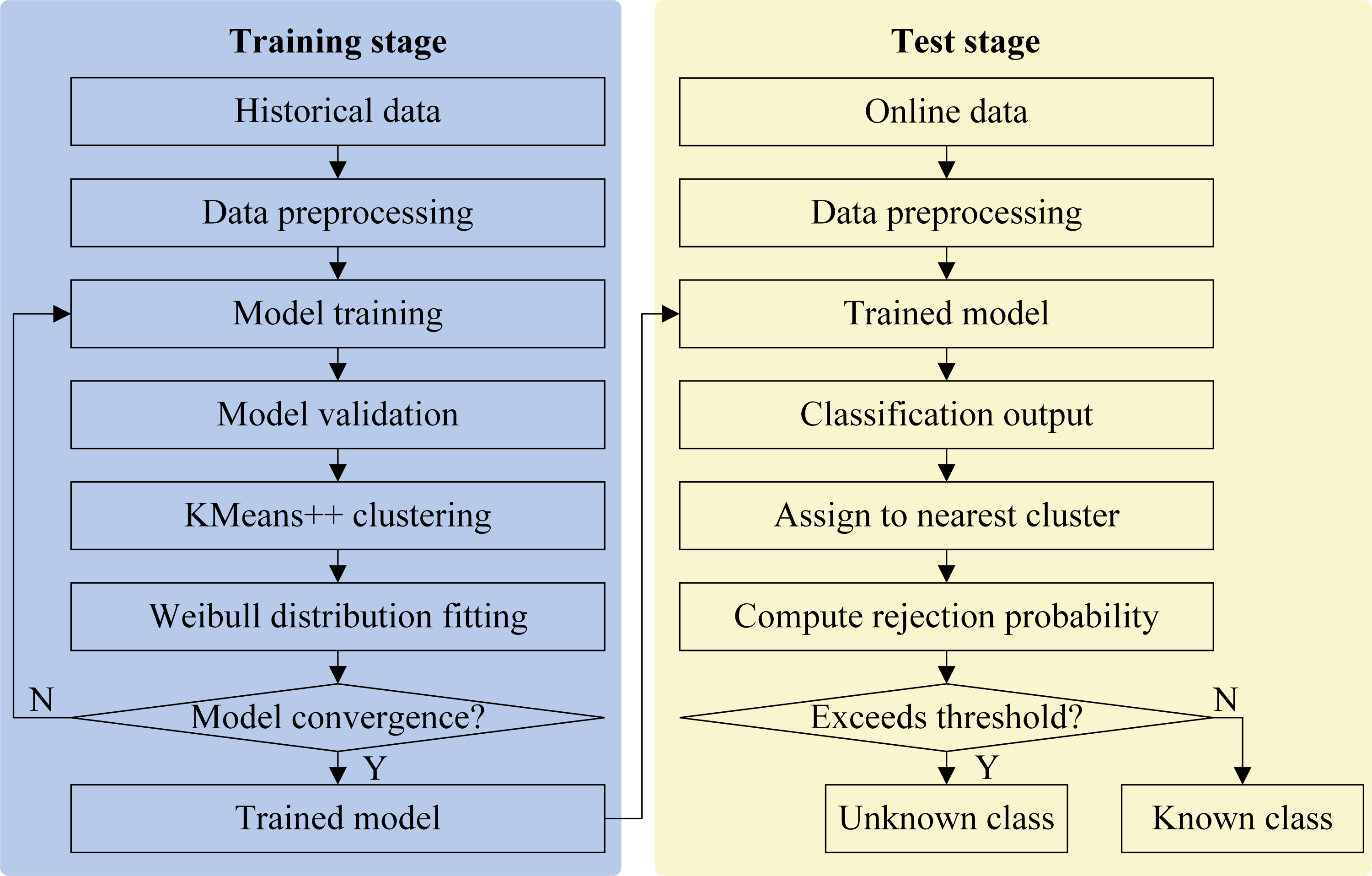}}
	\caption{Application workflow of FGCRN.}
	\label{Fig5}
\end{figure}

The application process of FGCRN is illustrated in \hyperref[Fig5]{Fig. 5}, which is partitioned into training and test stages. 

During the training stage, historical data are first standardized using z-score transformation to eliminate scale differences among variables. 
Then, the model’s feature extractor and classifier are optimized by minimizing training loss (see \hyperref[Eq15]{Eq. 15}). 
Model performance is evaluated using both the training and validation datasets. 
Subsequently, fine-grained feature representations are constructed. 
Correctly classified samples in each health state are selected and clustered into $M$ groups using the K-means++ model. 
The tail distribution of Mahalanobis distances between samples and their corresponding cluster centers is modeled using the Weibull distribution. 
This process is iteratively repeated until model convergence.

During the test stage, online data are standardized using statistical parameters derived from historical data. Then, the standardized data are input into the trained model to obtain classification results. Subsequently, each test sample is allocated to the cluster within its predicted category that has the smallest Mahalanobis distance. 
Finally, the rejection probability of the sample belonging to the assigned cluster is calculated using \hyperref[Eq12]{Eq. 12}. 
If the sample’s rejection probability exceeds a predefined threshold, it is identified as the unknown fault; otherwise, it is assigned a label according on the classification output.

\section{Experimental study}
\label{sec4}
\subsection{Evaluation metrics}

To clearly describe the evaluation metrics for open-set fault diagnosis, confusion matrix for category $Y_c$ is provided in \hyperref[Table1]{Table 1}. 
$TP_c$, $TN_c$, $FP_c$, and $FN_c$ denote the number of known-category samples correctly classified as $Y_c$, correctly classified as $Y_{c-}$, misclassified as $Y_c$, and misclassified as $Y_{c-}$, respectively. 
The number of samples correctly identified as the known category is calculated as $TK=TP_c +TN_c +FP_c +FN_c$. 
Additionally, $TU$, $FK$, and $FU$ denote the number of samples correctly identified as unknown category, misidentified as known category, and misidentified as unknown category, respectively.

\begin{table}[!ht]
\centering
	\label{Table1}
		\caption{Confusion matrix.}
\begin{threeparttable}
\begin{tabular}{lccc}
\hline
                          & PC is $Y_c$ & PC is $Y_{c-}$ & PC is unknown \\ \hline
RC is $Y_c$    & $TP_c$                        & $FN_c$                           & \multirow{2}{*}{$FU$}         \\
RC is $Y_{c-}$ & $FP_c$                        & $TN_c$                           &                             \\
RC is unknown  & \multicolumn{2}{c}{$FK$}                                   & $TU$                          \\ \hline
\end{tabular}
    \begin{tablenotes}
    \footnotesize
      \item[a] Note: RC denotes the real category, PC denotes the predicted category, $Y_{c-}$ denotes the known categories that are not $Y_{c}$.
    \end{tablenotes}
\end{threeparttable}

\end{table}

To comprehensively assess the classification performance for known health states and the identification capability for unknown faults, the accuracy for open set identification is adopted, formulated as follows,
\begin{equation}
	\begin{aligned}
    ACC=\frac{\sum_{c=1}^kTP_{c}+TU}{\sum_{c=1}^k\left(TP_{c}+FN_{c}\right)+TU+FK+FU}, 
    \end{aligned}
    \label{Eq16}	
\end{equation}

Additionally, misidentifying known health states as unknown fault may increase the workload of field operators, while misidentifying unknown fault as known health states could lead to incorrect decisions. 
To quantify these errors, the false acceptance rate (FAR) and false rejection rate (FRR) are introduced, which are formulated as follows,
\begin{equation}
	\begin{aligned}
    FAR&=\frac{FK}{FK+TU}, \\
    FRR&=\frac{FU}{\sum_{c=1}^k\left(TP_{c}+FN_{c}\right)+FU}. 
    \end{aligned}
    \label{Eq17}	
\end{equation}
The $FAR$ denotes the proportion of unknown faults that are incorrectly identified as known health states, while the $FRR$ refers to the proportion of known health states that are mistakenly identified as unknown faults. 
Higher $ACC$, lower $FAR$, and lower $FRR$ indicate better model performance.

\subsection{Implementation details}

The Adam optimizer was employed to update the weights of the FGCRN over 50 training epochs. 
The learning rate was initialized at 0.01 and decayed gradually according to the following schedule $ 0.01 \times 0.3^ {epoch//3}$. 
The batch size was 512, and the GRU hidden layer size was 100. 

Several advanced methods were selected for comparison, including MSP  \cite{RN266833}, OpenMax \cite{RN266834}, MaxLogit \cite{RN266836}, KL Match \cite{RN266836}, GEN \cite{RN266837} and ViM \cite{RN266838}. 
These methods have been demonstrated as capable of effectively identifying out-of-distribution samples or unknown categories. 
In this study, these methods are employed to evaluate the challenges of identifying unknown faults in multimode processes and validate the effectiveness of the proposed model.

\subsection{TE process}

\begin{figure}[!ht]
\centerline{\includegraphics[width=1\columnwidth,]{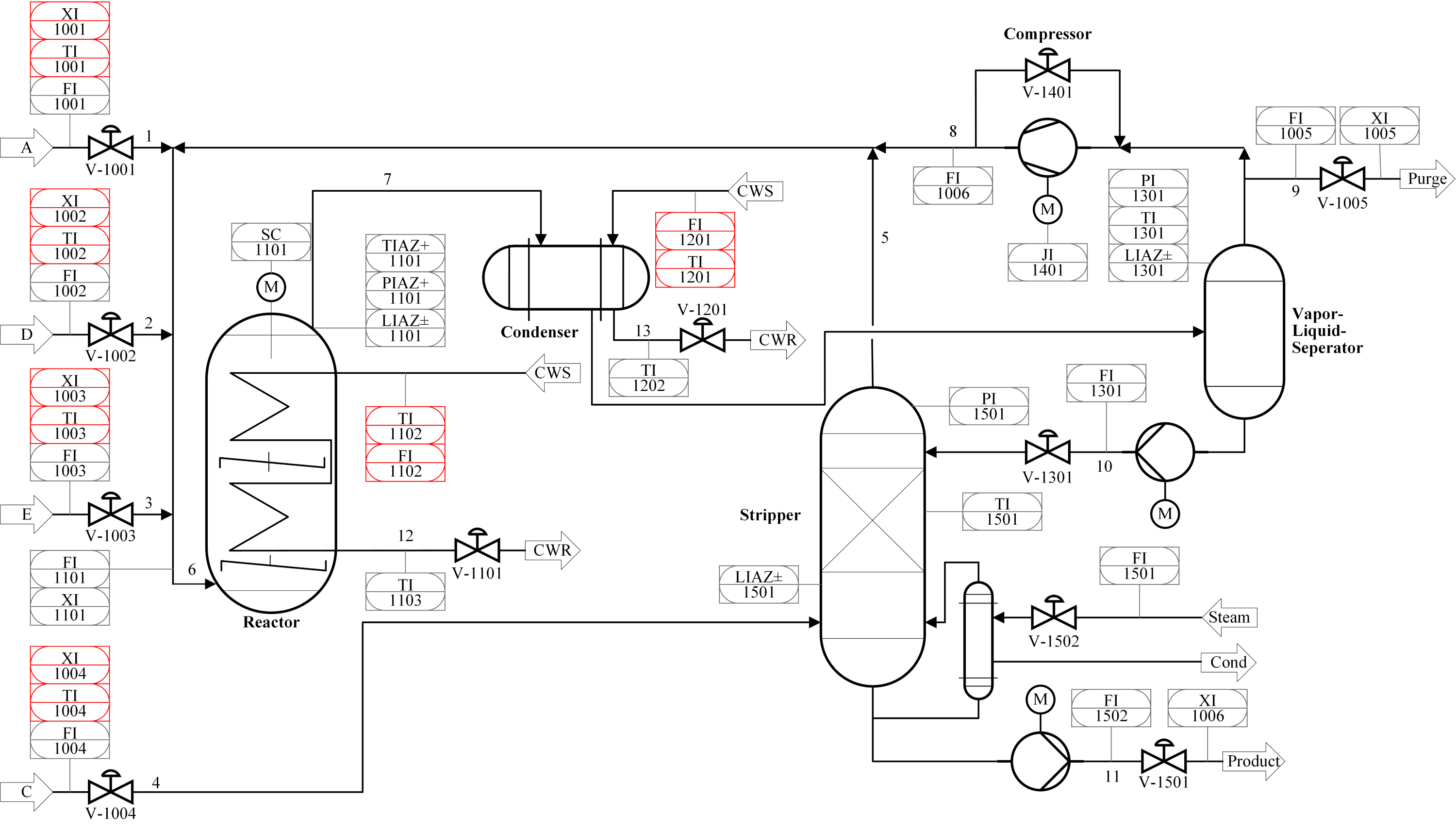}}
	\caption{P\&ID of the revised process model \cite{RN237021}.}
	\label{Fig6}
\end{figure}

TE process \cite{RN237020} is commonly employed to assess the performance of fault diagnosis models.  
This process has been implemented in Simulink \cite{RN237021}, and its corresponding system structure is shown in \hyperref[Fig6]{Fig. 6}. 
TE process includes six operating modes (see \hyperref[Table2]{Table 2}) and can simulate 28 types of faults.
Sixteen health state categories were selected for the experiment, as listed in \hyperref[Table3]{Table 3}. 
The task settings are listed in \hyperref[Table4]{Table 4}.
Six tasks were designed by combining different operating modes, where step fault F6, random variation fault F12, and unknown fault F20 were used as the unknown fault in subtasks A, B, and C, respectively. 
The experimental data were generated by TE process simulation platform, with a total simulation time of 100 hours, a sampling interval of 3 minutes, and fault injection occurring at the 30th hour. 
For the known health state samples, an 8:1:1 split was applied to generate training, validation, and test sets, whereas all unknown fault samples were used exclusively for test. 

\begin{table}[!ht]
	\centering
	\caption{Modes of TE process \cite{RN237020}.}
	\label{Table2}
\begin{tabular}{lll}
\hline
No. & G/H mass ratio & Production rate           \\ \hline
M1  & 50/50          & G: 7038kg/h, H: 7038kg/h  \\
M2  & 10/90          & G: 1408kg/h, H: 12669kg/h \\
M3  & 90/10          & G: 10000kg/h, H: 1111kg/h \\
M4  & 50/50          & maximum production rate   \\
M5  & 10/90          & maximum production rate   \\
M6  & 90/10          & maximum production rate   \\ \hline
\end{tabular}
\end{table}

\begin{table}[!ht]
	\centering
	\caption{Faults of TE process used in open set fault diagnosis \cite{RN237020}.}
	\label{Table3}
    \fontsize{10}{14}\selectfont
	\begin{tabular}{lll}
\hline
\textbf{No.} & \textbf{Description}                                     & \textbf{Type}    \\ \hline
N            & Normal                                                   & -                \\
F1           & A/C feed ratio, B composition constant (stream 4)        & Step             \\
F2           & B composition, A/C ratio constant (Stream 4)             & Step             \\
F4           & Reactor cooling water inlet temperature                  & Step             \\
F6           & A feed loss (stream 1)                                   & Step             \\
F7           & C header pressure loss - reduced availability (stream 4) & Step             \\
F8           & A, B, C feed composition (stream 4)                      & Random variation \\
F10          & C feed temperature (stream 4)                            & Random variation \\
F11          & Reactor cooling water inlet temperature                  & Random variation \\
F12          & Condenser cooling water inlet temperature                & Random variation \\
F13          & Reaction kinetics                                        & Drift            \\
F14          & Reactor cooling water valve                              & Sticking         \\
F17-20       & Unknown                                                  & Unknown          \\ \hline
\end{tabular}
\end{table}

\begin{table}[!ht]
\centering
	\label{Table4}
		\caption{Task settings on TE process dataset.}
\begin{tabular}{lll}
\hline
\textbf{Modes} & \textbf{Known category}                         & \textbf{Unknown category (Task label)} \\ \hline
M1,M4          & N,F1,F2,F4,F7,F8,F10,F11,F13,F14,F17-F19 & F6 (T1A) / F12 (T1B) / F20   (T1C)  \\
M2,M5          & N,F1,F2,F4,F7,F8,F10,F11,F13,F14,F17-F19 & F6 (T2A) / F12 (T2B) / F20   (T2C)  \\
M3,M6          & N,F1,F2,F4,F7,F8,F10,F11,F13,F14,F17-F19 & F6 (T3A) / F12 (T3B) / F20   (T3C)  \\
M1,M2          & N,F1,F2,F4,F7,F8,F10,F11,F13,F14,F17-F19 & F6 (T4A) / F12 (T4B) / F20   (T4C)  \\
M3,M4          & N,F1,F2,F4,F7,F8,F10,F11,F13,F14,F17-F19 & F6 (T5A) / F12 (T5B) / F20   (T5C)  \\
M5,M6          & N,F1,F2,F4,F7,F8,F10,F11,F13,F14,F17-F19 & F6 (T6A) / F12 (T6B) / F20   (T6C)  \\ \hline
\end{tabular}
\end{table}

The comparison of diagnostic accuracy among different models is shown in \hyperref[Table5]{Table 5}. 
The proposed model achieves the highest accuracy across all 18 subtasks. 
The average accuracy of all comparison models does not exceed 83\%, indicating that existing methods still face significant challenges in open-set fault diagnosis for multimode processes.
As shown in \hyperref[Fig1]{ Fig. 1}, the feature distributions of different health states tend to cluster according to operating modes. 
Such complex data distributions make it difficult to accurately identify unknown faults using a single scoring function or category representation. 
Compared with MSP, OpenMax, MaxLogit, KL Match, GEN, and ViM, the proposed model improves average accuracy by 15.17\%, 24.88\%, 25.42\%, 19.46\%, 15.19\%, and 24.16\%, respectively. 
These results demonstrate its superior performance in open-set fault diagnosis for multimode processes.

\begin{table}[!ht]
\centering
	\label{Table5}
		\caption{Accuracy for open
set fault diagnosis on TE process dataset.}
\begin{tabular}{llllllll}
\hline
\textbf{}    & \textbf{MSP} & \textbf{OpenMax} & \textbf{MaxLogit} & \textbf{KL Match} & \textbf{GEN} & \textbf{ViM} & \textbf{Proposed} \\ \hline
\textbf{T1A} & 94.35\%      & 93.02\%          & 91.78\%           & 95.38\%          & 94.74\%       & 94.41\%      & \textbf{97.51\%}  \\
\textbf{T1B} & 95.48\%      & 69.81\%          & 69.94\%           & 84.43\%          & 92.12\%       & 58.85\%      & \textbf{97.85\%}  \\
\textbf{T1C} & 83.81\%      & 53.86\%          & 55.11\%           & 67.25\%          & 81.03\%       & 76.35\%      & \textbf{97.31\%}  \\
\textbf{T2A} & 94.84\%      & 93.43\%          & 92.40\%           & 94.58\%          & 95.12\%       & 94.81\%      & \textbf{98.05\%}  \\
\textbf{T2B} & 82.61\%      & 63.52\%          & 89.74\%           & 65.06\%          & 91.10\%       & 55.42\%      & \textbf{98.02\%}  \\
\textbf{T2C} & 63.66\%      & 57.10\%          & 57.73\%           & 60.05\%          & 62.20\%       & 55.50\%      & \textbf{98.42\%}  \\
\textbf{T3A} & 94.38\%      & 93.35\%          & 92.99\%           & 94.95\%          & 95.36\%       & 97.27\%      & \textbf{98.56\%}  \\
\textbf{T3B} & 79.72\%      & 88.82\%          & 56.15\%           & 88.54\%          & 86.96\%       & 95.71\%      & \textbf{99.05\%}  \\
\textbf{T3C} & 93.45\%      & 66.51\%          & 68.54\%           & 86.10\%          & 88.87\%       & 65.12\%      & \textbf{98.26\%}  \\
\textbf{T4A} & 91.93\%      & 91.80\%          & 91.14\%           & 92.19\%          & 91.86\%       & 93.87\%      & \textbf{98.19\%}  \\
\textbf{T4B} & 87.42\%      & 68.23\%          & 77.30\%           & 84.75\%          & 88.37\%       & 55.53\%      & \textbf{98.62\%}  \\
\textbf{T4C} & 56.86\%      & 56.69\%          & 56.38\%           & 56.72\%          & 56.80\%       & 62.98\%      & \textbf{98.09\%}  \\
\textbf{T5A} & 92.80\%      & 91.77\%          & 91.85\%           & 94.19\%          & 93.35\%       & 94.72\%      & \textbf{97.25\%}  \\
\textbf{T5B} & 72.33\%      & 53.50\%          & 53.75\%           & 60.74\%          & 67.78\%       & 57.03\%      & \textbf{98.22\%}  \\
\textbf{T5C} & 74.50\%      & 53.65\%          & 54.95\%           & 64.69\%          & 69.87\%       & 56.90\%      & \textbf{97.04\%}  \\
\textbf{T6A} & 94.58\%      & 94.09\%          & 92.90\%           & 93.75\%          & 94.14\%       & 96.90\%      & \textbf{98.30\%}  \\
\textbf{T6B} & 80.57\%      & 71.09\%          & 57.19\%           & 73.40\%          & 82.89\%       & 62.64\%      & \textbf{98.20\%}  \\
\textbf{T6C} & 58.26\%      & 56.54\%          & 57.20\%           & 57.53\%          & 58.63\%       & 55.70\%      & \textbf{97.66\%}  \\
\textbf{Avg} & 82.86\%      & 73.15\%          & 72.61\%           & 78.57\%          & 82.84\%       & 73.87\%      & \textbf{98.03\%}  \\ \hline
\end{tabular}
\end{table}

The FRR results are summarized in \hyperref[Table6]{Table 6}. 
All models maintain a low misidentification rate of known categories as unknown faults, with average FRRs between 1\% and 2\%. 
This demonstrates that these models are able to accurately identify known health state categories.
The proposed model achieves the lowest average FRR of 1.12\%, effectively reducing the workload of field personnel caused by unnecessary verification of unknown fault categories when known health states are mistakenly identified as unknown faults. 

\begin{table}[!ht]
\centering
	\label{Table6}
		\caption{FRR for open set fault diagnosis on TE process dataset.}
\begin{tabular}{llllllll}
\hline
\textbf{}    & \textbf{MSP} & \textbf{OpenMax} & \textbf{MaxLogit} & \textbf{KL Match} & \textbf{GEN} & \textbf{ViM} & \textbf{Proposed} \\ \hline
\textbf{T1A} & 2.28\%       & 2.43\%           & 2.13\%            & 2.22\%           & 2.28\%        & 1.13\%       & 1.15\%            \\
\textbf{T1B} & 2.28\%       & 2.43\%           & 2.13\%            & 2.22\%           & 2.28\%        & 1.13\%       & 1.26\%            \\
\textbf{T1C} & 2.28\%       & 2.43\%           & 2.13\%            & 2.22\%           & 2.28\%        & 1.13\%       & 1.26\%            \\
\textbf{T2A} & 1.40\%       & 1.24\%           & 1.37\%            & 1.24\%           & 1.32\%        & 1.21\%       & 1.26\%            \\
\textbf{T2B} & 1.40\%       & 1.24\%           & 1.37\%            & 1.24\%           & 1.32\%        & 1.21\%       & 1.15\%            \\
\textbf{T2C} & 1.40\%       & 1.24\%           & 1.37\%            & 1.24\%           & 1.32\%        & 1.21\%       & 1.15\%            \\
\textbf{T3A} & 1.26\%       & 0.85\%           & 1.13\%            & 1.15\%           & 1.15\%        & 1.18\%       & 1.15\%            \\
\textbf{T3B} & 1.26\%       & 0.85\%           & 1.13\%            & 1.15\%           & 1.15\%        & 1.18\%       & 0.93\%            \\
\textbf{T3C} & 1.26\%       & 0.85\%           & 1.13\%            & 1.15\%           & 1.15\%        & 1.18\%       & 0.93\%            \\
\textbf{T4A} & 1.90\%       & 1.46\%           & 2.01\%            & 1.59\%           & 1.98\%        & 1.13\%       & 0.93\%            \\
\textbf{T4B} & 1.90\%       & 1.46\%           & 2.01\%            & 1.59\%           & 1.98\%        & 1.13\%       & 1.15\%            \\
\textbf{T4C} & 1.90\%       & 1.46\%           & 2.01\%            & 1.59\%           & 1.98\%        & 1.13\%       & 1.15\%            \\
\textbf{T5A} & 2.31\%       & 1.42\%           & 2.34\%            & 2.40\%           & 2.28\%        & 1.10\%       & 1.15\%            \\
\textbf{T5B} & 2.31\%       & 1.42\%           & 2.34\%            & 2.40\%           & 2.28\%        & 1.10\%       & 1.07\%            \\
\textbf{T5C} & 2.31\%       & 1.42\%           & 2.34\%            & 2.40\%           & 2.28\%        & 1.10\%       & 1.07\%            \\
\textbf{T6A} & 1.54\%       & 1.43\%           & 1.40\%            & 1.62\%           & 1.48\%        & 1.10\%       & 1.07\%            \\
\textbf{T6B} & 1.54\%       & 1.43\%           & 1.40\%            & 1.62\%           & 1.48\%        & 1.10\%       & 1.15\%            \\
\textbf{T6C} & 1.54\%       & 1.43\%           & 1.40\%            & 1.62\%           & 1.48\%        & 1.10\%       & 1.15\%            \\
\textbf{Avg} & 1.78\%       & 1.47\%           & 1.73\%            & 1.70\%           & 1.75\%        & 1.14\%       & \textbf{1.12\%}   \\ \hline
\end{tabular}
\end{table}

\begin{figure}[!ht]
\centerline{\includegraphics[width=1\columnwidth,]{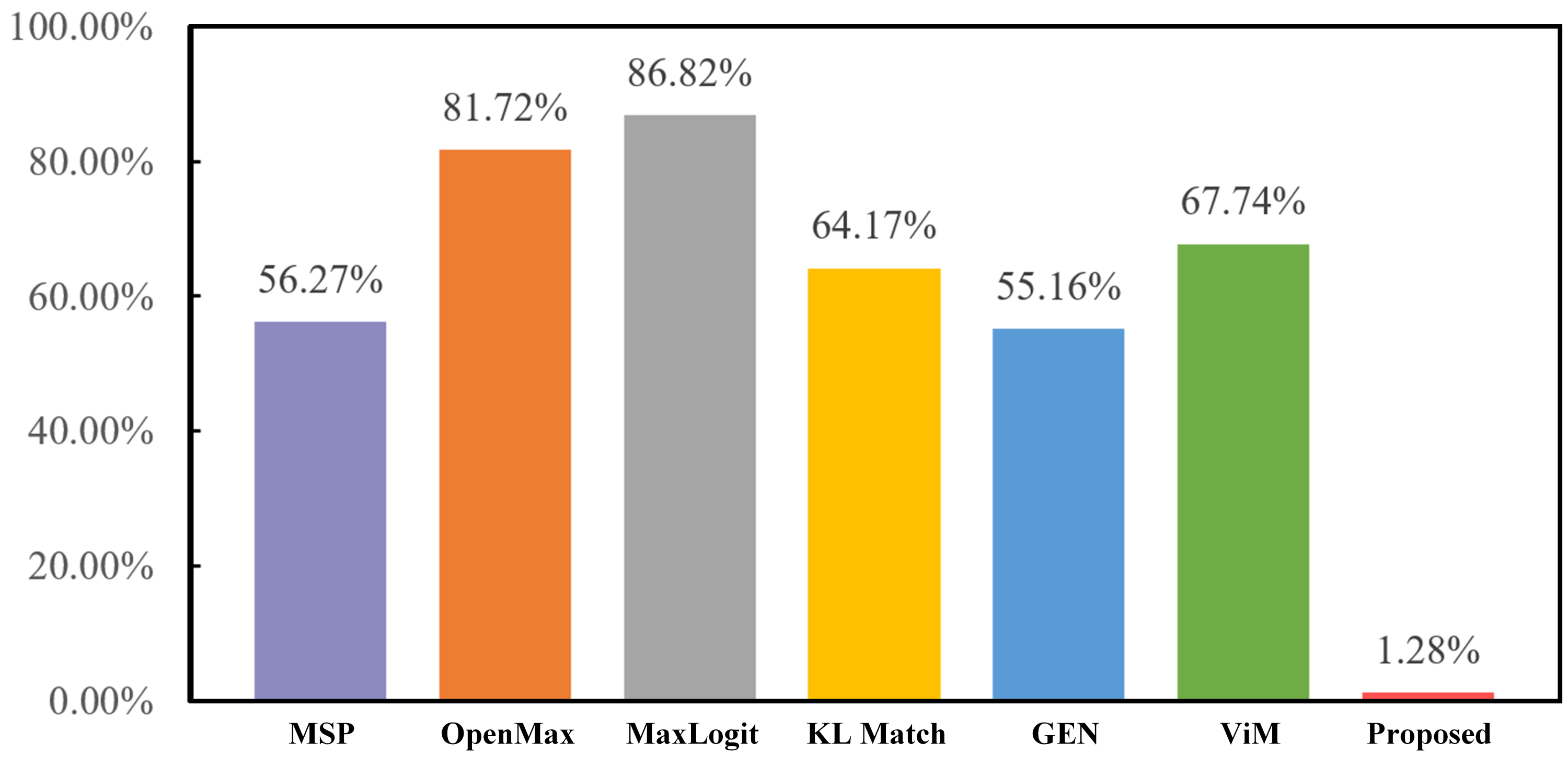}}
	\caption{Average FAR for open set fault diagnosis on TE process dataset.}
	\label{Fig7}
\end{figure}

The average FAR results are presented in \hyperref[Fig7]{ Fig. 7}. 
The comparison models show average FARs exceeding 50\%, indicating a high risk of misidentifying unknown faults as known health states. 
In contrast, the proposed model achieves a significantly lower FAR of only 1.28\%, effectively avoiding the misidentification of unknown faults and reducing the risk of incorrect decisions. 

\subsection{CSTR}

\begin{figure}[!ht]
\centerline{\includegraphics[width=0.6\columnwidth,]{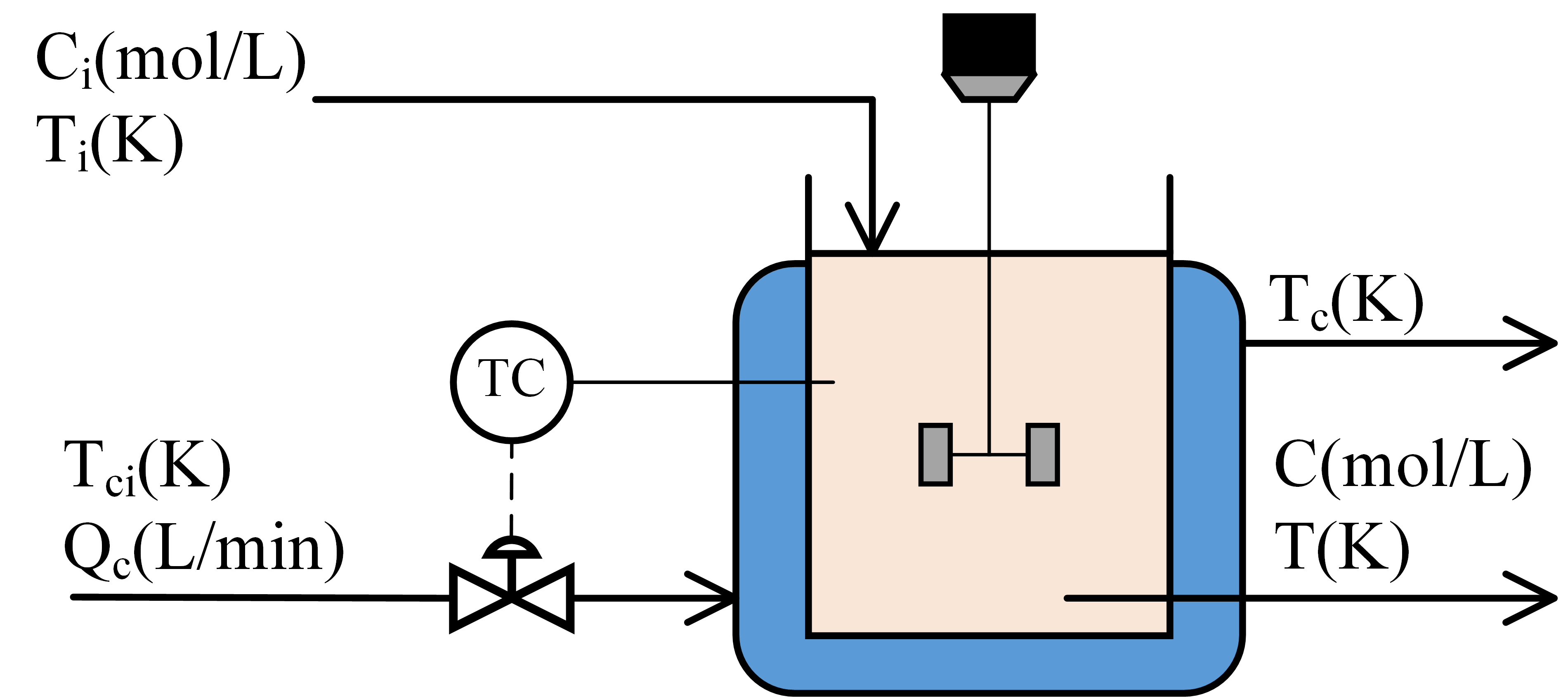}}
	\caption{Structure of CSTR \cite{RN237031}.}
	\label{Fig8}
\end{figure}

Continuous stirred tank reactor (CSTR) process \cite{RN237031} is also widely used to access fault diagnosis models, and its structure is presented in \hyperref[Fig8]{ Fig. 8}.
The control of reactor temperature $T$ is achieved through the regulation of cooling water flow rate $Q$. 
Nine fault types are simulated, as listed in \hyperref[Table7]{Table 7}. 
The original setpoint is defined as operating mode M1, with setpoints increased by 5 K and 10 K defined as modes M2 and M3, respectively.
The task settings are listed in \hyperref[Table8]{Table 8}. 
Each task consists of a pairwise combination of two operating modes, and faults F8 and F9 are set as unknown faults, corresponding to subtasks A and B, respectively. 
The experimental data were generated using the Simulink model of the CSTR process over a total of 20 hours, sampled once per minute, with fault injection starting at the 200th minute. 
The division of training, validation and test sets follows the same strategy used in TE process.

\begin{table}[!ht]
\centering
	\label{Table7}
		\caption{Faults of CSTR used in open set fault diagnosis\cite{RN237031}.}
\begin{tabular}{ll}
\hline
\textbf{Health state} & \textbf{Description}           \\ \hline
\textbf{N}                & Normal                        \\
\textbf{F1}               & $ C_i=C_{i,0}+0.001t $         \\
\textbf{F2}               & $ T_i=T_{i,0}+0.05t $          \\
\textbf{F3}               & $ C=C_0+0.001t $               \\
\textbf{F4}               & $ T=T_0+0.05t $                \\
\textbf{F5}               & $ Q_c=Q_{c,0}-0.1t $           \\
\textbf{F6}               & $ T_{ci}=T_{ci,0}+0.05t $      \\
\textbf{F7}               & $ T_c=T_{c,0}+0.05t $          \\
\textbf{F8}               & $ a=a_0 \text{exp}(-0.0005t) $ \\
\textbf{F9}               & $ b=b_0 \text{exp}(-0.001t) $  \\ \hline
\end{tabular}
\end{table}

\begin{table}[!ht]
\centering
	\label{Table8}
		\caption{Task settings on CSTR dataset.}
\begin{tabular}{lll}
\hline
\textbf{Modes} & \textbf{Known category}   & \textbf{Unknown category (Task label)} \\ \hline
M1,M2          & N,F1,F2,F3,F4,F5,F6,F7 & F8 (T7A) / F9 (T7B)                 \\
M1,M3          & N,F1,F2,F3,F4,F5,F6,F7 & F8 (T8A) / F9 (T8B)                 \\
M2,M3          & N,F1,F2,F3,F4,F5,F6,F7 & F8 (T9A) / F9 (T9B)                 \\ \hline
\end{tabular}
\end{table}

The diagnostic accuracies of different models are compared in \hyperref[Table9]{Table 9}. 
The proposed model consistently attains the top accuracies across all subtasks. 
Compared with MSP, OpenMax, MaxLogit, KL Match, GEN, and ViM, the proposed method shows average accuracy improvements of 40.51\%, 27.52\%, 45.54\%, 37.67\%, 40.52\% and 33.02\%, respectively. 
These results highlight the strong capability of the proposed model in handling open-set fault diagnosis for multimode processes.

\begin{table}[!ht]
\centering
	\label{Table9}
		\caption{Accuracy for open set fault diagnosis on CSTR dataset.}
\begin{tabular}{llllllll}
\hline
             & \textbf{MSP} & \textbf{OpenMax} & \textbf{MaxLogit} & \textbf{KL Match} & \textbf{GEN} & \textbf{ViM} & \textbf{Proposed} \\ \hline
\textbf{T1A} & 61.30\%      & 68.10\%          & 57.18\%           & 63.19\%           & 61.30\%      & 62.04\%      & \textbf{99.07\%}  \\
\textbf{T1B} & 44.03\%      & 98.61\%          & 44.03\%           & 44.31\%           & 44.03\%      & 87.92\%      & \textbf{99.07\%}  \\
\textbf{T2A} & 67.87\%      & 56.71\%          & 76.94\%           & 67.78\%           & 67.82\%      & 56.30\%      & \textbf{98.33\%}  \\
\textbf{T2B} & 54.21\%      & 75.69\%          & 44.03\%           & 69.86\%           & 54.21\%      & 67.04\%      & \textbf{98.56\%}  \\
\textbf{T3A} & 61.71\%      & 43.94\%          & 50.05\%           & 56.62\%           & 61.71\%      & 43.98\%      & \textbf{99.21\%}  \\
\textbf{T3B} & 57.27\%      & 81.30\%          & 43.98\%           & 61.67\%           & 57.27\%      & 74.03\%      & \textbf{95.19\%}  \\
\textbf{Avg} & 57.73\%      & 70.73\%          & 52.70\%           & 60.57\%           & 57.72\%      & 65.22\%      & \textbf{98.24\%}  \\ \hline
\end{tabular}
\end{table}

\begin{figure}[!ht]
\centerline{\includegraphics[width=1\columnwidth,]{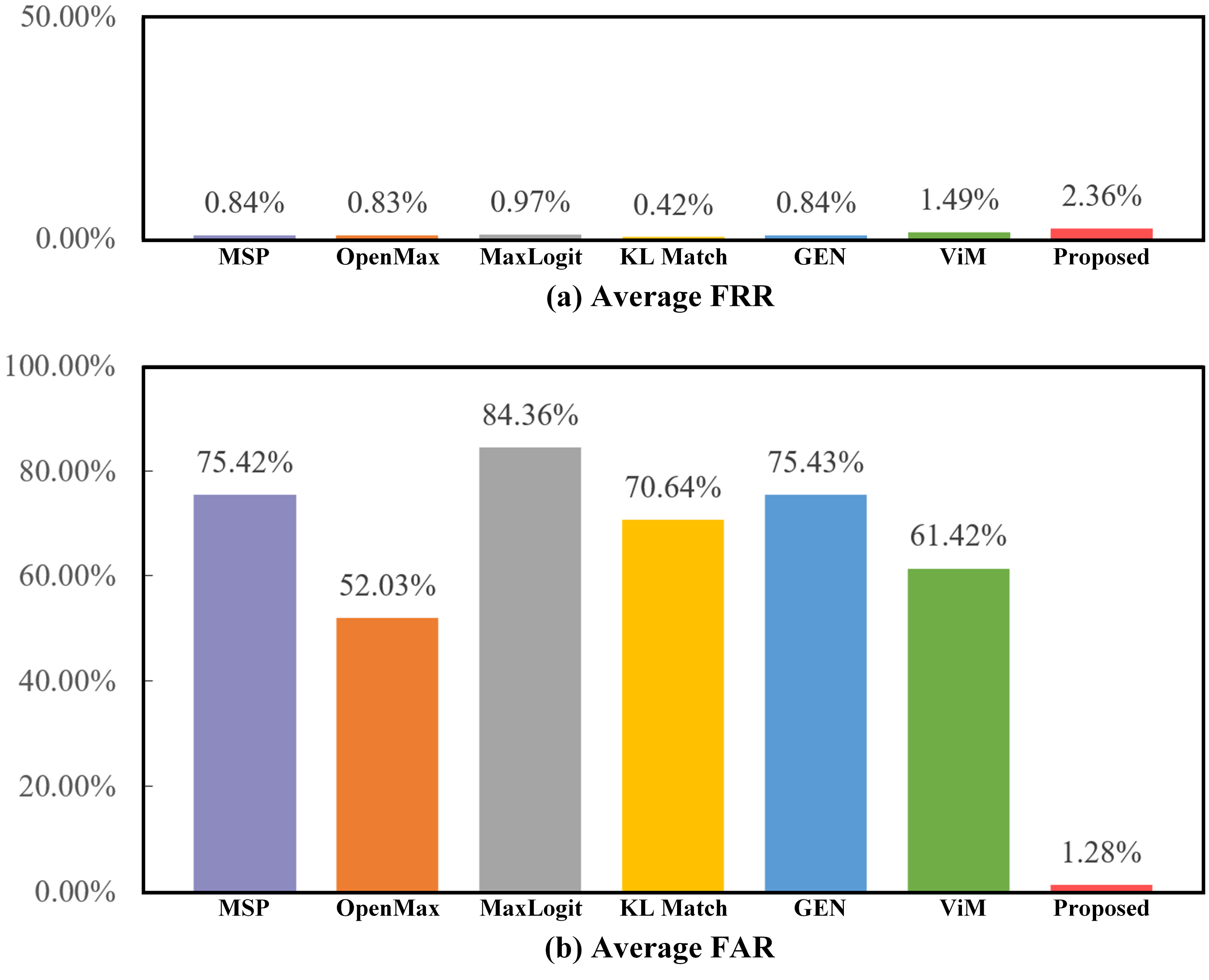}}
	\caption{Average FRR and FAR for open set fault diagnosis on CSTR dataset.}
	\label{Fig9}
\end{figure}

The average FRR and FAR of each model are shown in \hyperref[Fig9]{ Fig. 9}. 
While all models exhibit relatively low average FRRs, there are significant differences in average FARs. 
The proposed model achieves the lowest average FAR, indicating its strong capability in accurately identifying unknown faults, whereas the comparison models are more prone to misidentifying unknown faults as known health states.

\subsection{IPCTF}

\begin{figure}[!ht]
\centerline{\includegraphics[width=1\columnwidth,]{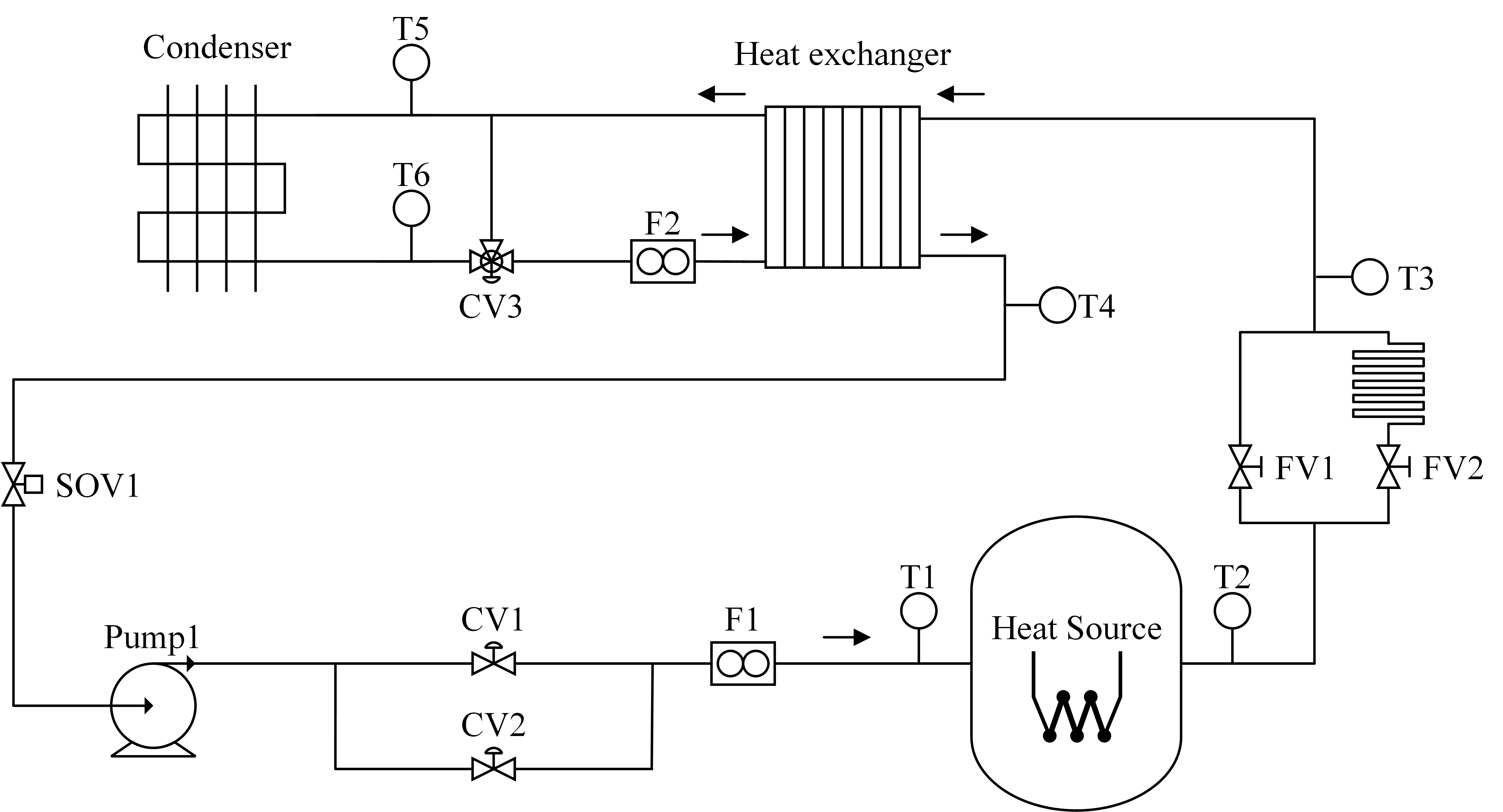}}
	\caption{Structure of IPCTF.}
	\label{Fig10}
\end{figure}

The intelligent process control-test facility developed by Wuhan University of Technology (see \hyperref[Fig10]{ Fig. 10}) was employed to evaluate the performance of the proposed model in the real system. 
This system controls the temperature of the heat source loop via the condenser. 
Experimental data were collected under three health states: normal (N), pump blockage in the heat source loop (F1), and pipeline blockage in the condenser loop (F2). 
The operating modes and task settings are listed in \hyperref[Table10]{Table 10}.

\begin{table}[!ht]
\centering
	\label{Table10}
		\caption{Operating
modes and task settings for open set fault diagnosis on IPCTF dataset.}
\begin{tabular}{lll}
\hline
\textbf{Mode No.} & \textbf{Power of the heat source (kW)} & \textbf{Setpoint of the cooling loop (℃)} \\ \hline
\textbf{M1}       & 10                                     & 29                                        \\
\textbf{M2}       & 8                                      & 24                                        \\ \hline
\textbf{Task No.} & \textbf{Known category}              & \textbf{Unknown category}                   \\ \hline
\textbf{T10A}     & N,F1                                   & F2                                        \\
\textbf{T10B}     & N,F2                                   & F1                                        \\ \hline
\end{tabular}
\end{table}

The diagnostic performance comparison among the evaluated models is shown in \hyperref[Table11]{Table 11}. 
The proposed method consistently outperforms the others in both subtasks. 
Moreover, the average accuracy of the proposed model and ViM significantly outperform all other comparison methods, demonstrating strong potential for practical system deployment.

\begin{table}[!ht]
\centering
	\label{Table11}
		\caption{Accuracy for open set fault diagnosis on IPCTF dataset.}
\begin{tabular}{llllllll}
\hline
              & \textbf{MSP} & \textbf{OpenMax} & \textbf{MaxLogit} & \textbf{KL Match} & \textbf{GEN} & \textbf{ViM} & \textbf{Proposed} \\ \hline
\textbf{T10A} & 28.97\%      & 35.72\%          & 28.97\%           & 28.97\%           & 28.97\%      & 99.51\%      & \textbf{99.72\%}  \\
\textbf{T10B} & 16.58\%      & 59.80\%          & 16.72\%           & 16.63\%           & 16.58\%      & 99.40\%      & \textbf{99.65\%}  \\
\textbf{Avg}  & 22.78\%      & 47.76\%          & 22.85\%           & 22.80\%           & 22.78\%      & 99.46\%      & \textbf{99.69\%}  \\ \hline
\end{tabular}
\end{table}

\begin{figure}[!ht]
\centerline{\includegraphics[width=1\columnwidth,]{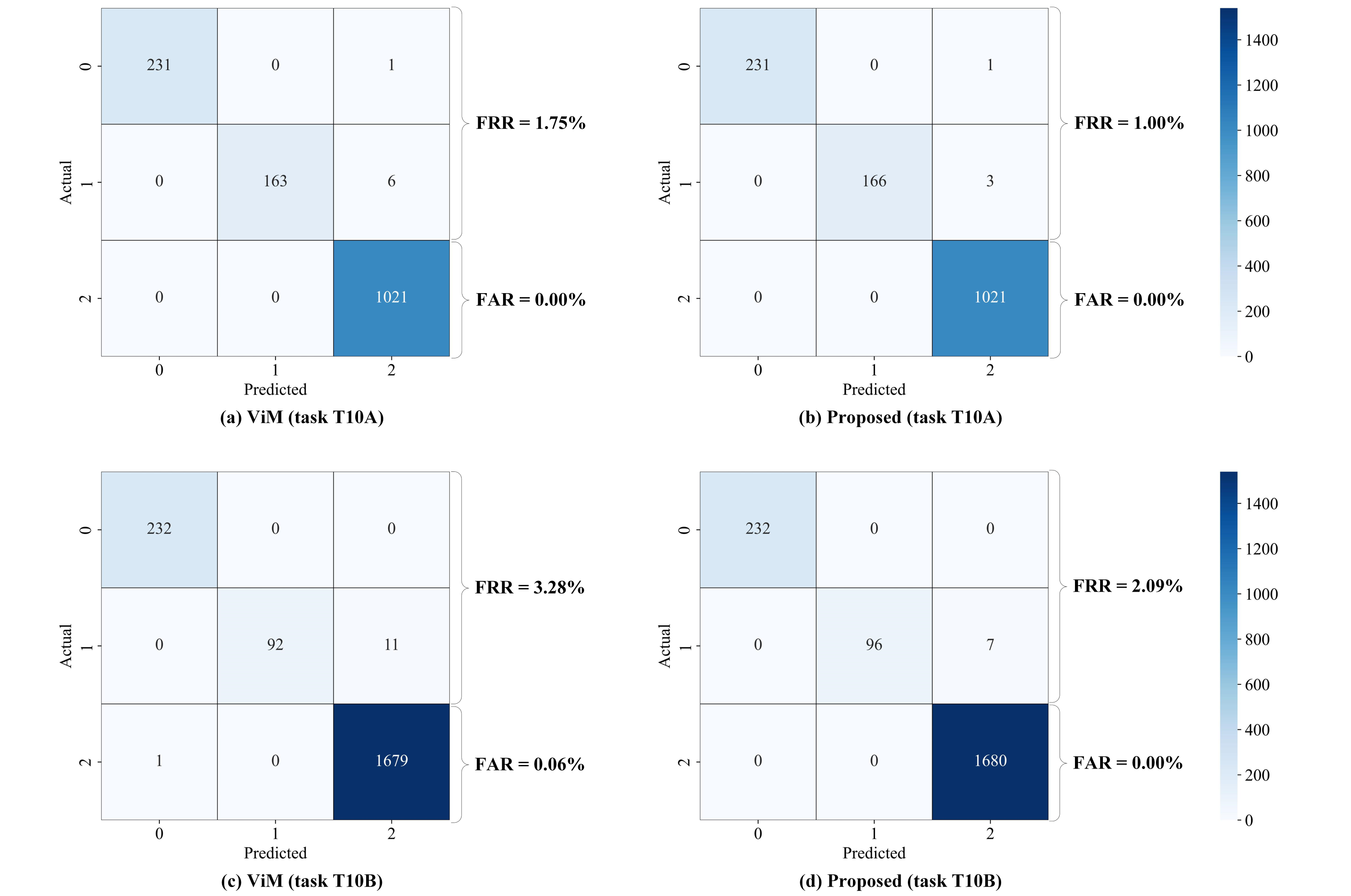}}
	\caption{Confusion matrix for open set fault diagnosis on IPCTF dataset.}
	\label{Fig11}
\end{figure}

The confusion matrices of our method and ViM are presented in \hyperref[Fig11]{ Fig. 11}. 
Labels 0 and 1 correspond to known health states, while label 2 denotes the unknown fault. 
In subtask T10A, both models effectively identify all unknown faults, with a FAR of 0\%. 
However, ViM exhibits a higher FRR than the proposed model. 
In subtask T10B, the proposed method outperforms ViM in both FAR and FRR, further demonstrating its practical effectiveness. 

\begin{figure}[!ht]
\centerline{\includegraphics[width=1\columnwidth,]{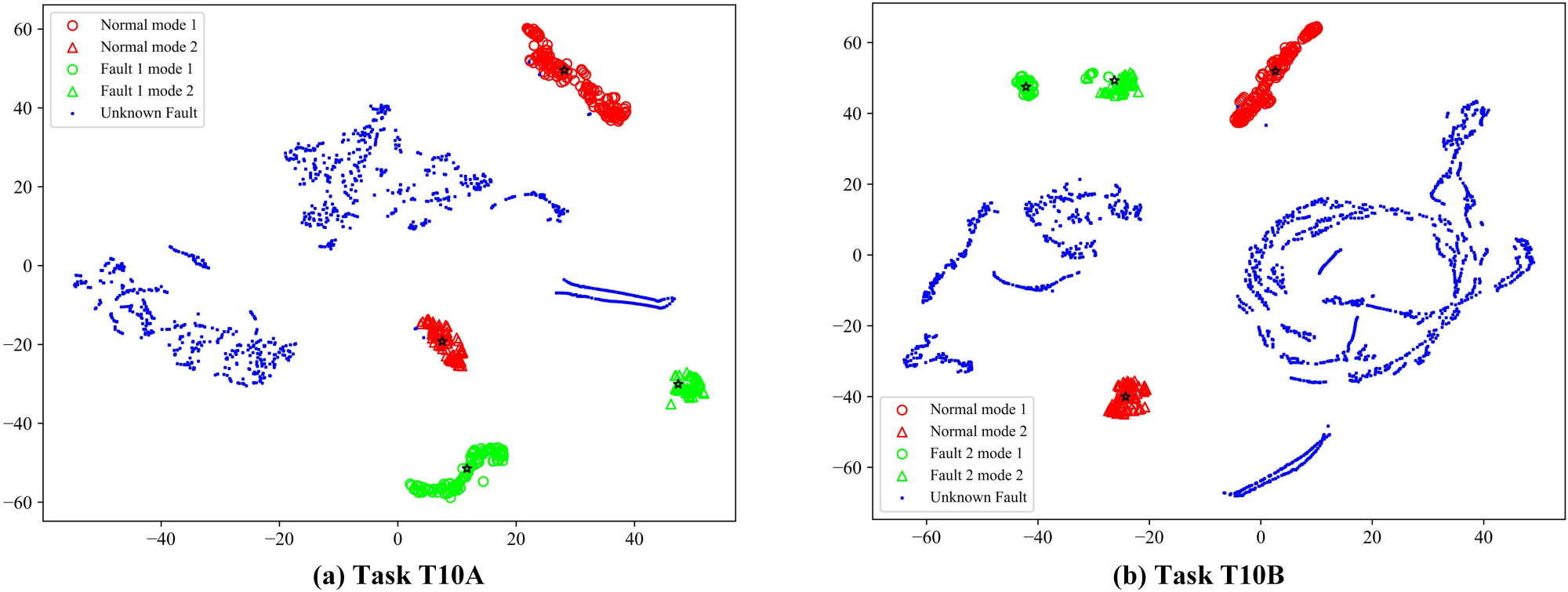}}
	\caption{Feature visualization on IPCTF dataset.}
	\label{Fig12}
\end{figure}

t-SNE was employed to visualize the feature representations generated by the proposed model, as illustrated in \hyperref[Fig12]{ Fig. 12}. 
In subtasks T10A and T10B, the features of normal samples exhibit clustered distributions corresponding to two operating modes, while the unknown-category samples are more dispersed in the feature space. 
This makes it challenging to construct a unified decision hyperplane that includes only normal samples. 
The pentagrams in the figure indicate the cluster centers. 
The results clearly demonstrate that the clusters associated with these centers are well-separated and highly distinguishable. 
This study quantifies the confidence that a sample belongs to the unknown category based on its Mahalanobis distance to cluster centers, effectively addressing the open-set fault diagnosis problem under multimode conditions.

\subsection{Ablation Study}

Ablation studies were carried out to examine the effectiveness of each component in the proposed method. 
The configurations of different model variants are listed in \hyperref[Table12]{Table 12}. 
Specifically, A1 replaces BiGRU with GRU; A2 removes the TAM module; A3 employs BN for normalization; A4 uses SAIN for normalization; A5 removes the $L_2$ loss term; and A6 constructs a single category representation for each health state. 

\begin{table}[!ht]
\centering
	\label{Table12}
		\caption{Configurations of different model variants for ablation experiments.}
\begin{threeparttable}
\begin{tabular}{ccccccc}
\hline
\textbf{}        & \textbf{A1}  & \textbf{A2}  & \textbf{A3}  & \textbf{A4}  & \textbf{A5}  & \textbf{A6}  \\ \hline
\textbf{BiGRU}   & GRU          & $\checkmark$ & $\checkmark$ & $\checkmark$ & $\checkmark$ & $\checkmark$ \\
\textbf{TAM}     & $\checkmark$ & -            & $\checkmark$ & $\checkmark$ & $\checkmark$ & $\checkmark$ \\
\textbf{BN+SAIN} & $\checkmark$ & $\checkmark$ & BN           & SAIN         & $\checkmark$ & $\checkmark$ \\
\textbf{$L_2$}      & $\checkmark$ & $\checkmark$ & $\checkmark$ & $\checkmark$ & -           & $\checkmark$ \\
\textbf{Cluster} & $\checkmark$ & $\checkmark$ & $\checkmark$ & $\checkmark$ & $\checkmark$ & -            \\ \hline
\end{tabular}
\end{threeparttable}
\end{table}

The diagnosis performance of different models is compared in \hyperref[Table13]{Table 13}. 
The proposed FGCRN obtains the top average accuracy of 98.03\%. 
Additionally, it attains the highest minimum accuracy across tasks at 96.81\%, outperforming models A1 to A6 by 12.65\%, 33.93\%, 4.05\%, 9.90\%, 33.11\%, and 40.52\%, respectively. 
These results indicate that the proposed method has strong identification capabilities for unknown faults under multiple operating modes, while demonstrating greater stability and robustness.
Compared with models A1 and A2, the proposed model achieves average classification accuracy increases of 2.27\% and 5.98\%, respectively. 
This reveals that the backward flow in BiGRU effectively captures health-related temporal features, and the TAM further enhances the model's discriminative ability by focusing on critical forward and backward time steps. 
Compared with A3 and A4, the proposed model shows improvements of 0.57\% and 2.08\%, respectively, which validates the complementarity of SAIN and BN. 
SAIN adaptively eliminates statistical interference, while BN preserves discriminative statistical features, thereby enhancing the model's reliability across multiple scenarios. 
For example, the accuracy of A3 on subtask T1B improved from 92.99\% to 97.85\% after incorporating the SAIN module. 
Similarly, the accuracy of A4 on subtask T2B increased from 87.14\% to 98.02\% after integrating the BN module. 
Furthermore, the proposed model outperformed A5 by 5.09\%, confirming that the Mahalanobis distance loss enhances feature compactness and aids in identifying unknown faults.  
It also outperformed A6 by 8.09\%, indicating that constructing fine-grained representations for each health state facilitates the formation of a reliable feature space constrained to corresponding category samples, thereby enhancing the identification of unknown faults. 
Overall, the combination of these components enables the proposed method to provide reliable and stable open-set fault diagnosis performance across multiple operating modes.

\begin{table}[!ht]
\centering
	\label{Table13}
		\caption{Accuracy for ablation experiments on TE process dataset.}

\begin{tabular}{llllllll}
\hline
\textbf{}    & \textbf{A1}      & \textbf{A2}      & \textbf{A3}      & \textbf{A4}      & \textbf{A5}      & \textbf{A6}      & \textbf{Proposed} \\ \hline
\textbf{T1A} & 93.10\%          & 94.07\%          & 96.98\%          & 95.51\%          & \textbf{97.51\%} & 97.12\%          & \textbf{97.51\%}  \\
\textbf{T1B} & 84.39\%          & 77.24\%          & 92.99\%          & 96.97\%          & 97.72\%          & \textbf{98.33\%} & 97.85\%           \\
\textbf{T1C} & 95.77\%          & 96.73\%          & 97.46\%          & 96.13\%          & 97.07\%          & \textbf{97.51\%} & 97.31\%           \\
\textbf{T2A} & 97.41\%          & 98.43\%          & 97.56\%          & \textbf{98.82\%} & 98.59\%          & 94.84\%          & 98.05\%           \\
\textbf{T2B} & 98.18\%          & \textbf{98.26\%} & 98.23\%          & 87.14\%          & 77.13\%          & 64.35\%          & 98.02\%           \\
\textbf{T2C} & 97.30\%          & 63.11\%          & 97.34\%          & 97.97\%          & 63.93\%          & 56.52\%          & \textbf{98.42\%}  \\
\textbf{T3A} & 98.38\%          & 97.22\%          & 97.91\%          & 98.74\%          & 99.30\%          & \textbf{99.05\%} & 98.56\%           \\
\textbf{T3B} & 98.87\%          & 98.94\%          & 98.63\%          & 99.11\%          & 99.49\%          & \textbf{99.49\%} & 99.05\%           \\
\textbf{T3C} & 97.95\%          & 97.98\%          & 97.50\%          & 97.95\%          & \textbf{98.52\%} & 96.48\%          & 98.26\%           \\
\textbf{T4A} & 97.22\%          & 96.22\%          & 97.68\%          & \textbf{98.34\%} & 96.45\%          & 93.41\%          & 98.19\%           \\
\textbf{T4B} & \textbf{98.70\%} & 98.46\%          & 98.32\%          & 98.25\%          & 93.32\%          & 95.59\%          & 98.62\%           \\
\textbf{T4C} & 88.80\%          & 77.20\%          & 97.72\%          & \textbf{98.18\%} & 82.48\%          & 65.30\%          & 98.09\%           \\
\textbf{T5A} & 96.80\%          & 94.94\%          & 96.97\%          & 92.21\%          & \textbf{97.86\%} & 96.64\%          & 97.25\%           \\
\textbf{T5B} & 95.66\%          & 92.80\%          & 97.93\%          & 87.45\%          & 85.49\%          & 82.42\%          & \textbf{98.22\%}  \\
\textbf{T5C} & 96.28\%          & 96.62\%          & 96.57\%          & 94.20\%          & 93.59\%          & 94.79\%          & \textbf{97.04\%}  \\
\textbf{T6A} & 97.96\%          & 96.80\%          & 98.27\%          & 97.62\%          & 98.92\%          & \textbf{98.94\%} & 98.30\%           \\
\textbf{T6B} & 93.59\%          & 92.70\%          & \textbf{98.68\%} & 97.56\%          & 97.42\%          & 90.06\%          & 98.20\%           \\
\textbf{T6C} & 97.34\%          & 89.21\%          & 97.59\%          & 95.03\%          & 98.06\%          & \textbf{98.04\%} & 97.66\%           \\
\textbf{Min} & 84.39\%          & 63.11\%          & 92.99\%          & 87.14\%          & 63.93\%          & 56.52\%          & \textbf{97.04\%}  \\
\textbf{Avg} & 95.76\%          & 92.05\%          & 97.46\%          & 95.95\%          & 92.94\%          & 89.94\%          & \textbf{98.03\%}  \\ \hline
\end{tabular}
\end{table}

\section{Conclusion}
\label{sec5}
This paper proposes a novel fine-grained clustering and rejection network for open-set fault diagnosis in multimode industrial processes. 
This method focuses on constructing multiple feature representations for each known health state to enhance the identification of unknown faults. 
Extensive experiments validate that the proposed method outperforms other advanced models. 
The principal conclusions are outlined as follows. 
\begin{itemize}
\item The integration of MSDC, BiGRU, and TAM enables efficient extraction of deep feature representations associated with known health states from complex data distributions.

\item The combination of BN and SAIN adaptively preserves discriminative statistical features, enhancing the model's adaptability to diverse operating modes. 

\item Distance loss enhances the discriminability of the feature space by compressing the feature distributions within the same health state, thereby facilitating the identification of unknown faults based on feature distance.

\item Constructing fine-grained representations for each health state further depicts more detailed internal structures, which significantly strengthens the model’s capability to identify unknown faults.
\end{itemize}

Although this study has achieved encouraging results, there are still some limitations. 
The proposed model categorizes all unknown faults into a single "unknown" category without performing a more detailed unsupervised classification. 
Moreover, once these unknown faults are labeled and incorporated into known categories, they could be utilized to further refine the diagnostic model. 
Future work could focus on developing automated incremental learning-based fault diagnosis models for multimode processes.


\bibliographystyle{elsarticle-num} 
\bibliography{ref}
\end{document}